\documentclass{article}

\usepackage{microtype}
\usepackage{graphicx}
\usepackage{subcaption}
\usepackage{booktabs}

\usepackage[english]{babel}
\usepackage[autolanguage]{numprint}
\nprounddigits{3}
\usepackage{multirow} 
\usepackage{siunitx}
\usepackage{enumitem}
\usepackage{ifthen}

\usepackage{hyperref}

\usepackage[accepted]{icml2025}

\usepackage{float}
\usepackage{textpos}
\usepackage{colortbl}
\usepackage{pifont}
\usepackage{xcolor}

\usepackage{etoolbox}
\usepackage{xparse}
\usepackage{tikz}
\usepackage{pgfplots}
\pgfplotsset{
    compat=1.17,
    compat/show suggested version=false,
    legend image code/.code={
        \draw[mark indices={2}] plot coordinates {
            (0cm,0cm) (0.15cm,0cm) (0.3cm,0cm)
        };
    }
}
\usepgfplotslibrary{fillbetween}

\definecolor{C0}{HTML}{0090FF}
\definecolor{C1}{HTML}{E54D2E}
\definecolor{C2}{HTML}{46A758}
\definecolor{C3}{HTML}{F76B15}
\definecolor{C4}{HTML}{D6409F}
\definecolor{C5}{HTML}{AD7F58}

\definecolor{customblue}{HTML}{2a6ca6}
\definecolor{customgreen}{HTML}{419136}
\definecolor{customred}{HTML}{c4352b}
\definecolor{custompurple}{HTML}{8361af}

\usepackage{amsmath}
\usepackage{amssymb}
\usepackage{mathtools}
\usepackage{amsthm}
\usepackage{bm, nicefrac}
\usepackage{xcolor}

\usepackage[capitalize,noabbrev]{cleveref}

\newcommand{\lft}{\mathopen{}\mathclose\bgroup\left}
\newcommand{\rgt}{\aftergroup\egroup\right}

\newcommand{\R}{\mathbb{R}}

\renewcommand{\vec}[1]{\bm{#1}}
\newcommand{\mat}[1]{\bm{#1}}

\newcommand{\transpose}[1]{#1^\top}

\NewDocumentCommand{\Gauss}{O{0}O{1}o}{\mathcal{N}\lft(#1,#2 \IfValueT{#3}{\mid #3}\rgt)}

\makeatletter
\DeclareRobustCommand\onedot{\futurelet\@let@token\@onedot}
\def\@onedot{\ifx\@let@token.\else.\null\fi\xspace}

\makeatother

\let\marginnotetemp\marginnote
\let\marginnote\relax
\usepackage{marginnote}
\let\marginnotepkg\marginnote

\NewDocumentCommand{\margintag}{O{0\baselineskip}m}{
    \checkoddpage
    \ifoddpage
        {\marginnotepkg{\footnotesize #2}[#1]}
    \else
        {\reversemarginpar\marginnotepkg{\footnotesize #2}[#1]}
    \fi}

\renewcommand\marginnote\marginnotetemp

\theoremstyle{plain}
\newtheorem{theorem}{Theorem}[section]

\newtheorem{lemma}[theorem]{Lemma}

\theoremstyle{definition}

\theoremstyle{remark}

\icmltitlerunning{JEDI: The Force of Jensen-Shannon Divergence in Disentangling Diffusion Models}

\begin{document}

\twocolumn[
\icmltitle{JEDI\\The Force of Jensen-Shannon Divergence in Disentangling Diffusion Models}



\icmlsetsymbol{equal}{*}

\begin{icmlauthorlist}
\icmlauthor{Eric Tillmann Bill}{eth}
\icmlauthor{Enis Simsar}{eth}
\icmlauthor{Thomas Hofmann}{eth}

\end{icmlauthorlist}

\icmlaffiliation{eth}{Department of Computer Science, ETH Zurich, Switzerland}
\icmlcorrespondingauthor{Eric Tillmann Bill}{erbill@ethz.ch}
\icmlkeywords{Image Generation, Diffusion Models, Test-Time Adaptation, Subject Disentanglement}

\vskip 0.3in
]



\printAffiliationsAndNotice{}  

\begin{abstract}
We introduce JEDI, a test-time adaptation method that enhances subject separation and compositional alignment in diffusion models without requiring retraining or external supervision. JEDI operates by minimizing semantic entanglement in attention maps using a novel Jensen-Shannon divergence based objective. To improve efficiency, we leverage adversarial optimization, reducing the number of updating steps required.
JEDI is model-agnostic and applicable to architectures such as Stable Diffusion 1.5 and 3.5, consistently improving prompt alignment and disentanglement in complex scenes. Additionally, JEDI provides a lightweight, CLIP-free disentanglement score derived from internal attention distributions, offering a principled benchmark for compositional alignment under test-time conditions. Code and results are available at \href{https://ericbill21.github.io/JEDI/}{ericbill21.github.io/JEDI/}.
\end{abstract}

\section{Introduction}\label{sec:intro}
Diffusion models have achieved remarkable success in generative modeling, particularly in the domain of image synthesis \citep{ho_denoising_2020, rombach_high-resolution_2022, lipman2022flow}. Among these, text-to-image (T2I) diffusion models \citep{esser_scaling_2024, ramesh_hierarchical_2022, podell_sdxl_2023} stand out for their ability to generate diverse and high-quality images conditioned on natural language prompts.

\begin{figure}[t]
    \centering
    \begin{subfigure}[t]{0.18\textwidth}
        \centering
        \captionsetup{labelformat=empty}
        \caption{\textbf{Stable Diffusion 3.5}}
        \includegraphics[width=\linewidth]{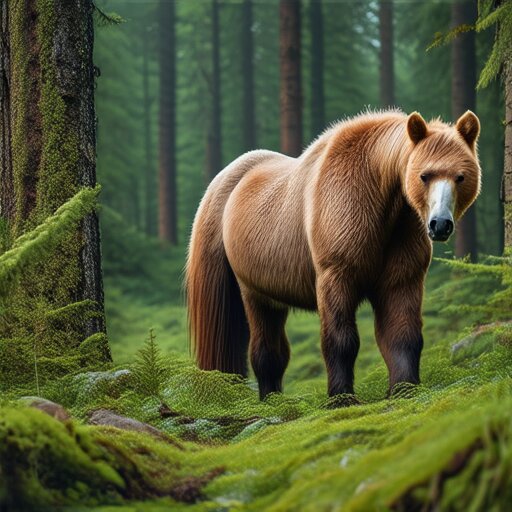}
    \end{subfigure}
    \hspace{8mm}
    \begin{subfigure}[t]{0.18\textwidth}
        \centering
        \captionsetup{labelformat=empty}
        \caption{\parbox{1.2\textwidth}{\hspace*{-0.12\textwidth}\textbf{Stable Diffusion 3.5 + JEDI}}}
        \includegraphics[width=\linewidth]{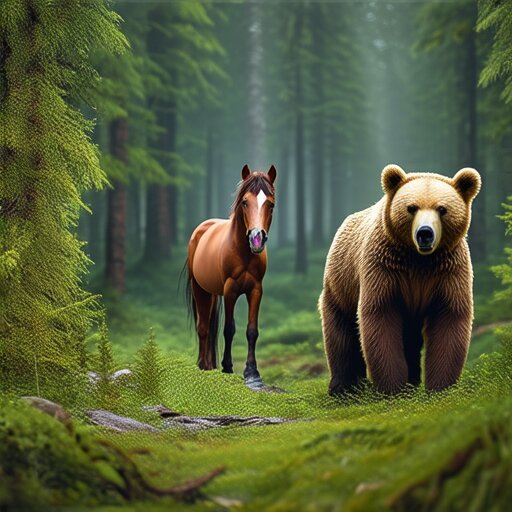}
    \end{subfigure}    
    \caption{\textbf{JEDI enables test-time subject disentanglement.} For the prompt \textit{``A \textbf{horse} and a \textbf{bear} in a forest''}, JEDI reduces attribute mixing and improves subject separation in Stable Diffusion 3.5.}
    \label{fig:intro_example}
    \vspace{-2mm}
\end{figure}

However, despite these advances, current T2I models often struggle with compositional prompts that involve multiple objects or intricate spatial relationships. For example, when given a prompt like \textit{``A horse and a bear in a forest,''} models from the Stable Diffusion family may produce semantically inconsistent outputs: one subject may be omitted (missing object), features from both animals may blend together into a single entity (attribute mixing), or the spatial arrangement may appear incoherent, refer to \cref{fig:intro_example}.

Such failures are especially problematic at \textit{test time}, where retraining or fine-tuning is often infeasible. To address these limitations, a range of test-time adaptation techniques have been proposed, which broadly fall into two categories: 1.) \textit{Latent Optimization Methods}, which adjust the latent representations during sampling to better align with the prompt \citep{meral_conform_2023, chefer_attend-and-excite_2023, wei_enhancing_2024}. 2.) \textit{Concept-Based Methods} which rely on external structural cues such as layouts or segmentation maps to guide the generation process \citep{kwon_concept_2024, binyamin_make_2024,liu_compositional_2023}.

While concept-based methods provide structural guidance, they often require additional models and can alter the underlying generative distribution. In contrast, latent optimization methods operate entirely within the model's architecture and offer a lightweight, model-preserving alternative for test-time adaptation.
In this work, we focus on latent optimization and introduce a novel, training-free test-time adaptation method called \textbf{JEDI} (\textbf{Je}nsen-Shannon Divergence for \textbf{D}isentanglement at \textbf{I}nference). By framing compositional entanglement as a probabilistic alignment problem, we propose a new divergence-based objective tailored for attention distributions. Our main contributions are as follows:\vspace{-2mm}

\begin{enumerate}[label=\roman*)] 
    \item We introduce a novel objective based on Jensen-Shannon divergence to minimize semantic entanglement in attention maps at test-time, providing a probabilistically grounded alternative to cosine similarity.
    \item By leveraging adversarial optimization techniques, we reduce the number of optimization steps, making JEDI lightweight and efficient for real-world use.
    \item JEDI demonstrates strong performance across multiple architectures, including Stable Diffusion 1.5, LoRACLR, and Stable Diffusion 3.5, consistently improving alignment with complex prompts.
    \item JEDI provides an entanglement score derived from internal attention maps, enabling compositional evaluation without relying on external models such as CLIP.
\end{enumerate}
\section{Latent Optimization}
Latent alignment methods steer the iterative denoising process in diffusion models by modifying the latent image during sampling. These methods often leverage model’s internal attention maps, which act as soft spatial probability distributions, indicating how strongly each token (e.g., \textit{``horse'', ``bear''}) influences different image regions.

At each timestep $t$ during inference, we retrieve the updated latent $\vec{x}_{t+1}$ and the internal attention maps $A_{t+1}$:
\begin{align*}
    \vec{x}_{t+1}, A_{t+1} = \mathrm{model}(\vec{x}_t, t).
\end{align*}
We then perform a test-time update of $\vec{x}_t$ by minimizing a disentanglement loss defined over $A_{t+1}$:
\begin{align*}
    \vec{x}_t \leftarrow \vec{x}_t - \alpha \nabla_{\vec{x}_t} \mathrm{score}(A_t),
\end{align*}
where $\mathrm{score}(A_t)$ penalizes overlap between attention maps of different entities. This encourages spatial disentanglement and mitigates attribute mixing. See \Cref{alg:jedi} in \Cref{sec:algo} for a pseudo-code implementation.

\noindent\textbf{Probabilistic View.}
Although attention maps are often treated as similarity scores, the use of the $\mathrm{softmax}$ function ensures that they are normalized and can instead be interpreted as discrete probability distributions.

Prior work \citep{meral_conform_2023, wei_enhancing_2024} overlooked this probabilistic structure, commonly relying on cosine similarity as a measure for alignment, despite its lack of probabilistic grounding. An exception is \citet{chefer_attend-and-excite_2023}, which considers attention probabilities but focuses only on maximizing individual token activation without accounting for inter-token competition.

In contrast, throughout this work we interpret attention maps as discrete probability distributions and optimize them accordingly. Our objective is to encourage unimodal, spatially localized, and non-overlapping attention for each subject in the prompt. This enables more faithful and disentangled representations, all achieved via test-time adaptation.

\section{Methodology}\label{sec:method}
We propose JEDI (\textbf{Je}nsen-Shannon Divergence for \textbf{D}isentanglement at \textbf{I}nference), a test-time adaptation method that improves subject separation in diffusion models by adjusting latent representations using attention statistics. Our objective combines Jensen-Shannon divergence (JSD) and Shannon Entropy to encourage intra-group coherence, inter-group separation, and spatial diversity.

\noindent\textbf{Jensen-Shannon Divergence.}
To measure the overlap among a set $P = \{\vec{p}_1, \dots, \vec{p}_n\} \subset \R^d$ of spatial attention distributions, we use the Jensen-Shannon divergence:
\begin{align*}
    D_\mathrm{JS}(P) = \frac{1}{|P|}\sum_{\vec{p} \in P} D_\mathrm{KL}\lft(\vec{p} \;\|\; \vec{m}\rgt), \quad \vec{m} = \frac{1}{|P|}\sum_{\vec{p} \in P} \vec{p},
\end{align*}
where $D_\mathrm{KL}$ is the Kullback-Leibler divergence, defined as:
\begin{align*}
    D_\mathrm{KL}(\vec{p} \;\|\; \vec{q}) = \sum_{i=1}^d p_i \log \frac{p_i}{q_i}.
\end{align*}

Since $D_\mathrm{JS}(P) \in [0, \log n]$ is bounded, we normalize it by dividing by $\log n$, yielding $\hat{D}_\mathrm{JS}(P) \in [0, 1]$, which enables comparison across groups of different sizes. For a formal proof of this bound, see \cref{lem:jsd} in \cref{sec:proofs}.

\begin{figure}[t]
    \centering
    \vspace{-2mm}
    \begin{subfigure}[t]{\linewidth}
        \centering
        \captionsetup{labelformat=empty}
        \caption{JEDI}
        \vspace{-1mm}
        \includegraphics[width=\textwidth]{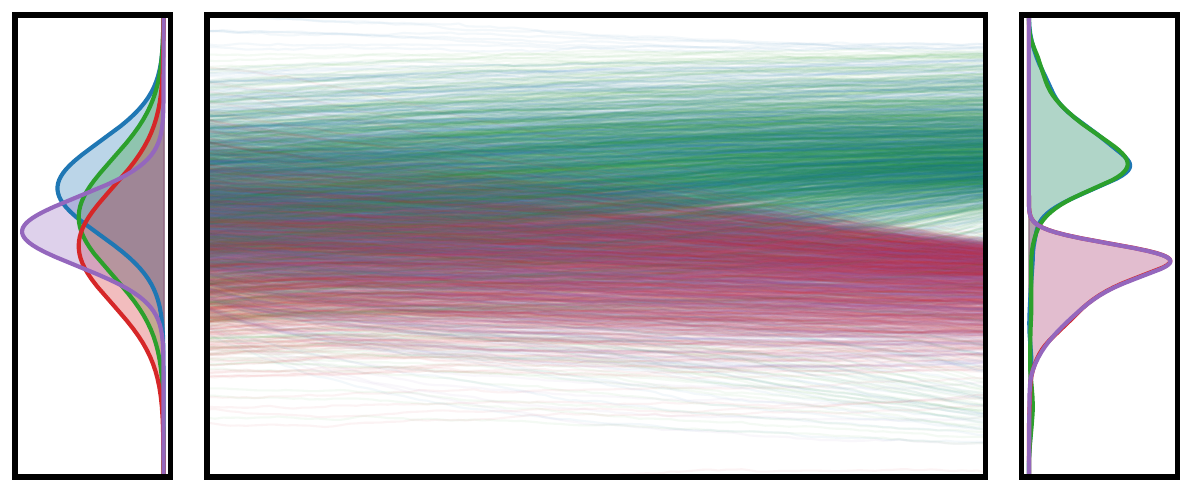}
    \end{subfigure}
    \\
    \vspace{-1mm}
    \begin{subfigure}[t]{\linewidth}
        \centering
        \captionsetup{labelformat=empty}
        \caption{NT-Xent Loss}
        \vspace{-1mm}
        \includegraphics[width=\linewidth]{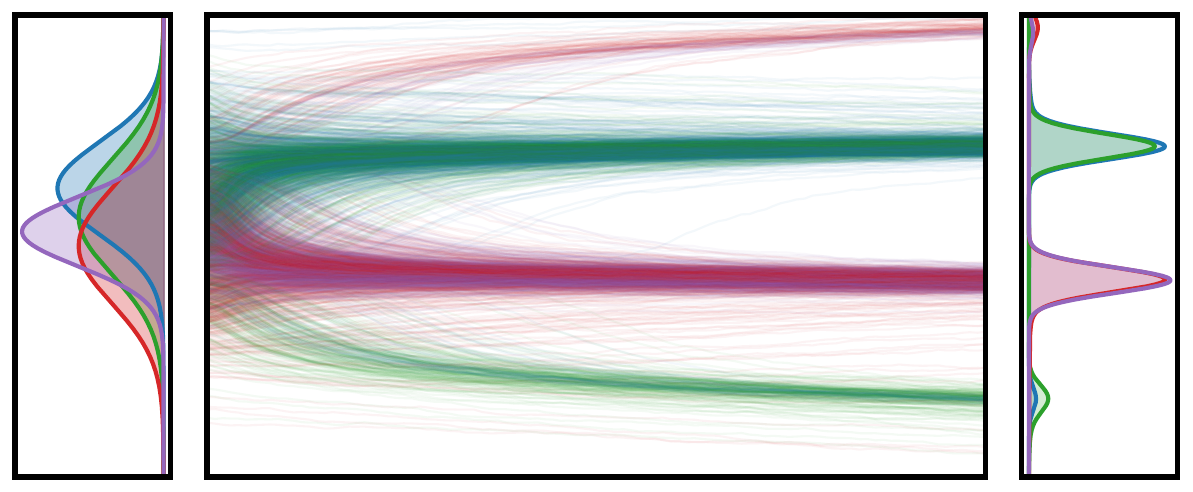}
    \end{subfigure}
    \caption{\textbf{Optimization Evolution of JEDI and NT-Xent.} Synthetic example with four overlapping distributions: \textcolor{customblue}{\textbf{blue}}/\textcolor{customgreen}{\textbf{green}} correspond to one subject, \textcolor{customred}{\textbf{red}}/\textcolor{custompurple}{\textbf{purple}} to another. Overlaps of \textcolor{customblue}{\textbf{blue}} and \textcolor{customgreen}{\textbf{green}} form teal, while \textcolor{customred}{\textbf{red}} and \textcolor{custompurple}{\textbf{purple}} form pink. JEDI preserves coherent group structure, while NT-Xent collapses modes.}
    \label{fig:ntx_vs_jsd}
    \vspace{-2mm}
\end{figure}

\begin{figure}
    \centering
    \begin{subfigure}[t]{0.3\linewidth}
        \centering
        \captionsetup{labelformat=empty}
        \caption{\textbf{SD 1.5}}
        \includegraphics[width=\linewidth]{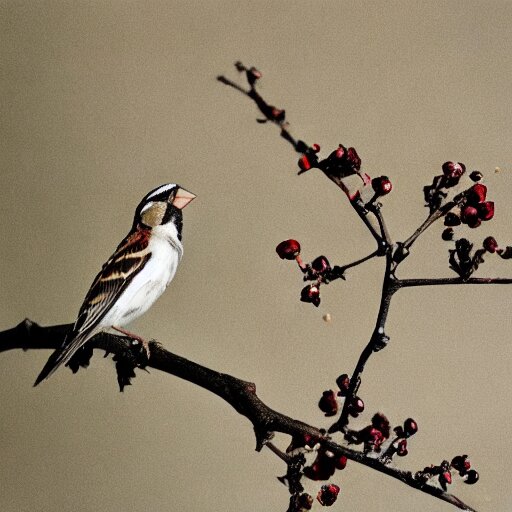}
    \end{subfigure}
    \begin{subfigure}[t]{0.3\linewidth}
        \centering
        \captionsetup{labelformat=empty}
        \caption{\textbf{JEDI (ours)}}
        \includegraphics[width=\linewidth]{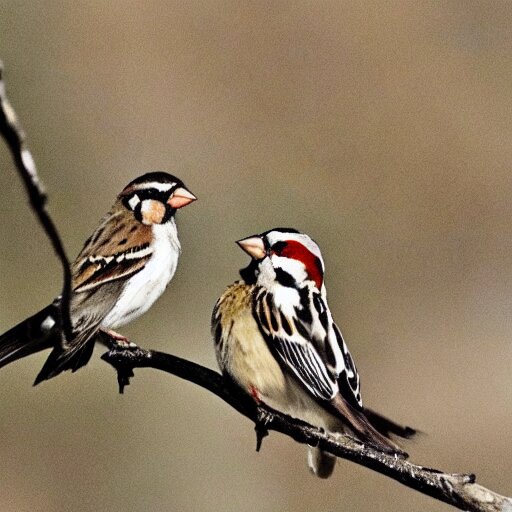}
    \end{subfigure}
    \begin{subfigure}[t]{0.3\linewidth}
        \centering
        \captionsetup{labelformat=empty}
        \caption{\textbf{CONFORM}}
        \includegraphics[width=\linewidth]{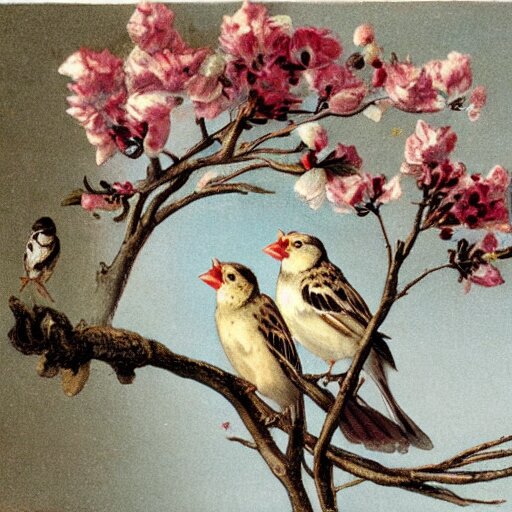}
    \end{subfigure}
    \par\vspace{0.5mm}
    {\small \textit{``A \textbf{sparrow} and a \textbf{finch} perched on a blossoming branch''}}
    \par\vspace{1.5mm}
    \begin{subfigure}[t]{0.3\linewidth}
        \centering
        \includegraphics[width=\linewidth]{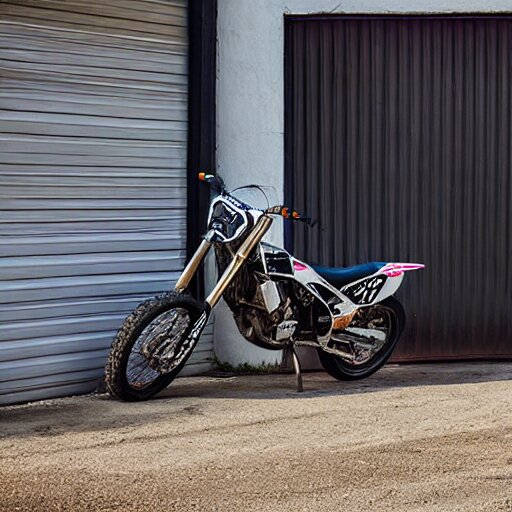}
    \end{subfigure}
    \begin{subfigure}[t]{0.3\linewidth}
        \centering
        \includegraphics[width=\linewidth]{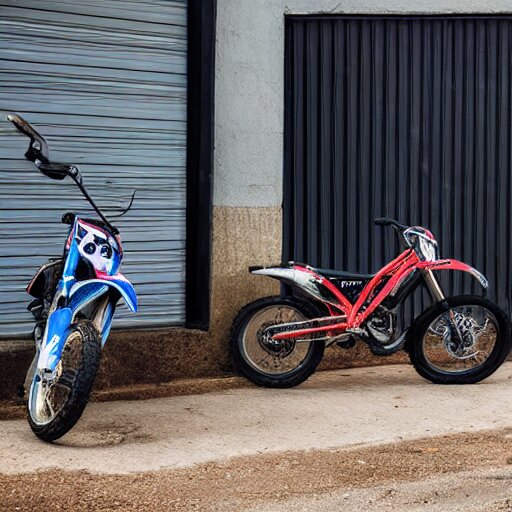}
    \end{subfigure}
    \begin{subfigure}[t]{0.3\linewidth}
        \centering
        \includegraphics[width=\linewidth]{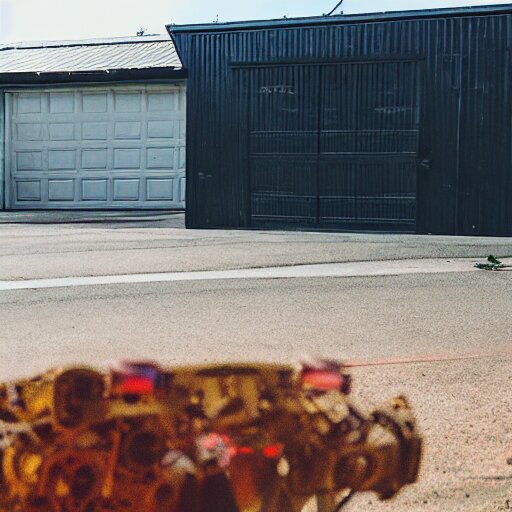}
    \end{subfigure}
    \par\vspace{-0.75mm}
    {\small \textit{``A \textbf{street bike} and a \textbf{dirt bike} leaning [...]''}}
    \caption{\textbf{Comparison of JEDI and CONFORM on Stable Diffusion 1.5.} Each image triplet was generated under identical conditions. For more details and examples, refer to \cref{sec:samples}.}
    \label{fig:conform}
\end{figure}

\noindent\textbf{Shannon Entropy.}
To control the sharpness of individual attention maps, we incorporate the Shannon entropy:
\begin{align*}
    H(\vec{p}) = - \sum_{i=1}^d p_i \log p_i.
\end{align*}
Entropy ranges from $0$ (single peak) to $\log d$ (uniform). We normalize it by $\log d$, yielding $\hat{H}(\vec{p}) \in [0, 1]$, which allows scale-independent balancing, where high entropy indicates spatial spread; low entropy implies tight localization. For a proof of the bound refer to \cref{lem:entropy} in \cref{sec:proofs}.

\begin{figure*}[t]
    \centering
    \begin{subfigure}[b]{0.32\textwidth}
        \centering
        {\small \textit{``A \textbf{dachshund} and a \textbf{corgi} sitting [...]''}}
        \begin{subfigure}[t]{0.48\textwidth}
            \centering
            \includegraphics[width=\linewidth]{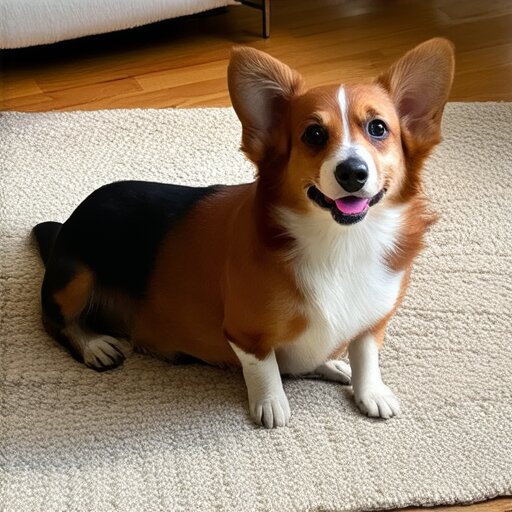}
        \end{subfigure}
        \hfill
        \begin{subfigure}[t]{0.48\textwidth}
            \centering
            \includegraphics[width=\linewidth]{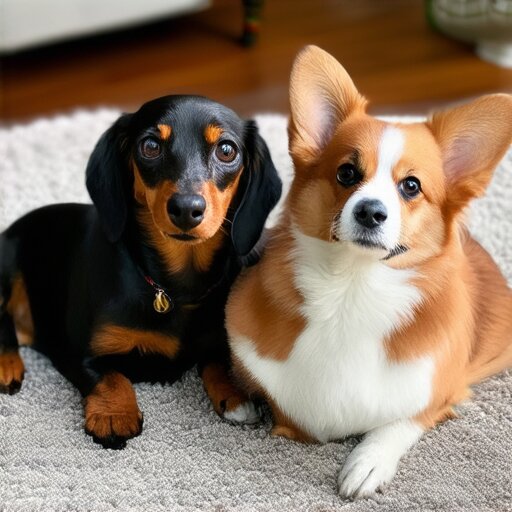}
        \end{subfigure}
    \end{subfigure}
    \hfill
    \begin{subfigure}[b]{0.32\textwidth}
        \centering
        {\small \textit{``A \textbf{street bike} and a \textbf{dirt bike} leaning [...]''}}
        \begin{subfigure}[t]{0.48\textwidth}
            \centering
            \includegraphics[width=\linewidth]{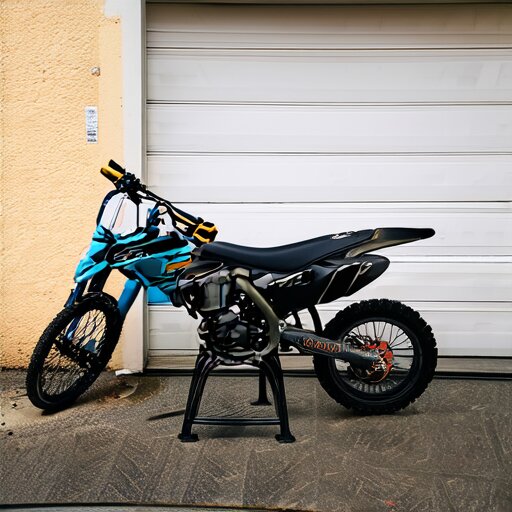}
        \end{subfigure}
        \hfill
        \begin{subfigure}[t]{0.48\textwidth}
            \centering
            \includegraphics[width=\linewidth]{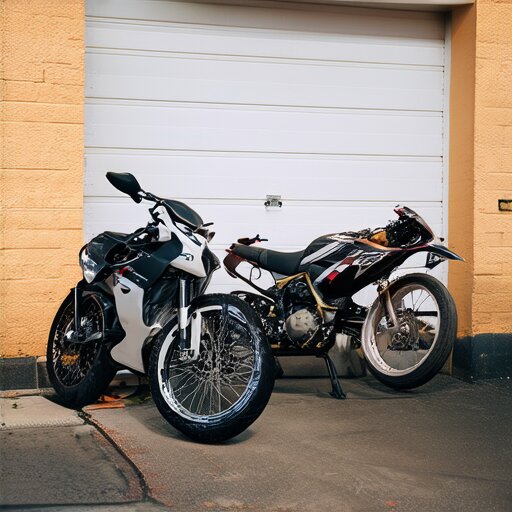}
        \end{subfigure}
    \end{subfigure}
    \hfill
    \begin{subfigure}[b]{0.32\textwidth}
        \centering
        {\small \textit{``A \textbf{sparrow} and a \textbf{finch} perched on [...]''}}
        \begin{subfigure}[t]{0.48\textwidth}
            \centering
            \includegraphics[width=\linewidth]{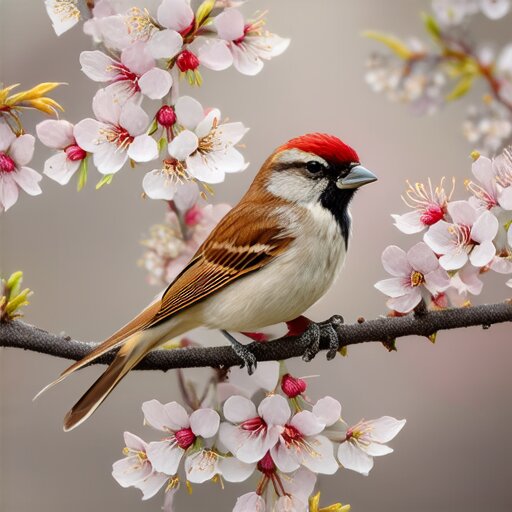}
        \end{subfigure}
        \hfill
        \begin{subfigure}[t]{0.48\textwidth}
            \centering
            \includegraphics[width=\linewidth]{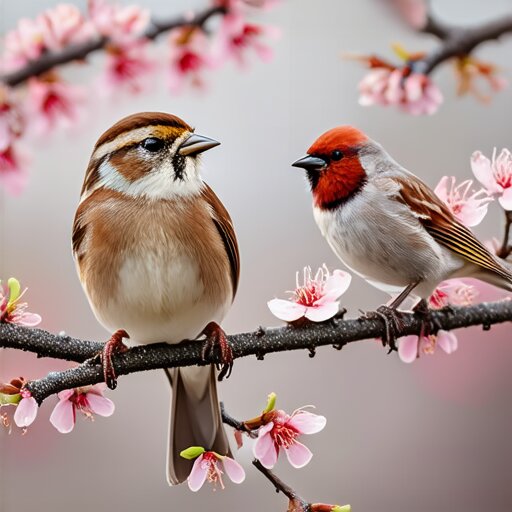}
        \end{subfigure}
    \end{subfigure}

    \vspace{1em}
    
    \begin{subfigure}[b]{0.32\textwidth}
        \centering
        {\small \textit{``A \textbf{black cat}, an \textbf{orange cat}, and\\a \textbf{white cat} lounging on a windowsill''}} \\
        \begin{subfigure}[t]{0.48\textwidth}
            \centering
            \includegraphics[width=\linewidth]{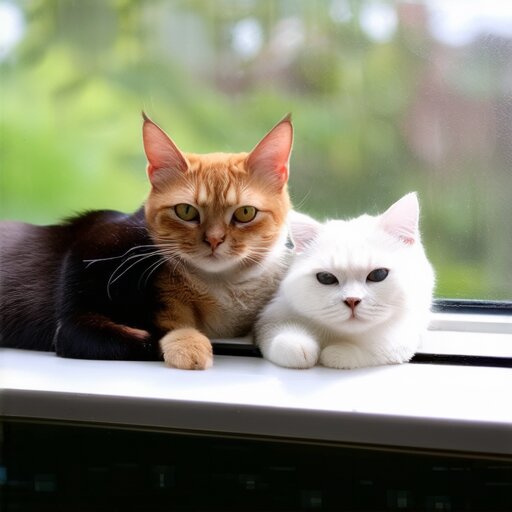}
        \end{subfigure}
        \hfill
        \begin{subfigure}[t]{0.48\textwidth}
            \centering
            \includegraphics[width=\linewidth]{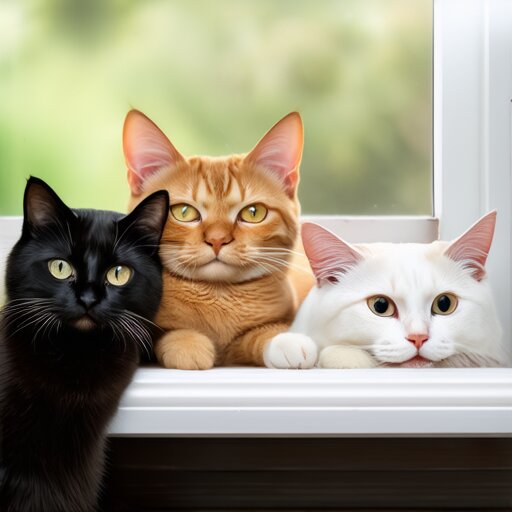}
        \end{subfigure}
    \end{subfigure}
    \hfill
    \begin{subfigure}[b]{0.32\textwidth}
        \centering
        {\small \textit{``A \textbf{horse}, a \textbf{bear}, and\\a \textbf{moose} in a forest clearing''}}\\
        \begin{subfigure}[t]{0.48\textwidth}
            \centering
            \includegraphics[width=\linewidth]{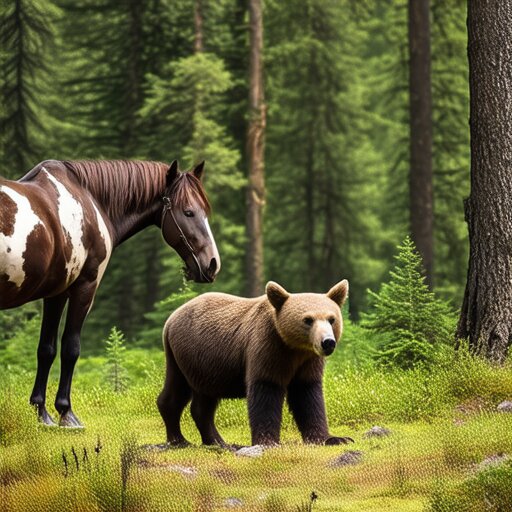}
        \end{subfigure}
        \hfill
        \begin{subfigure}[t]{0.48\textwidth}
            \centering
            \includegraphics[width=\linewidth]{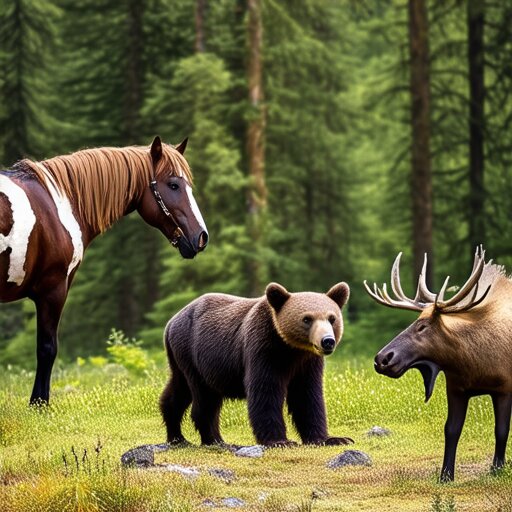}
        \end{subfigure}
    \end{subfigure}
    \hfill
    \begin{subfigure}[b]{0.32\textwidth}
        \centering
         {\small \textit{``A \textbf{Labrador}, a \textbf{Golden Retriever}, and a \textbf{German Shepherd} playing in a backyard''}}
        \begin{subfigure}[t]{0.48\textwidth}
            \centering
            \includegraphics[width=\linewidth]{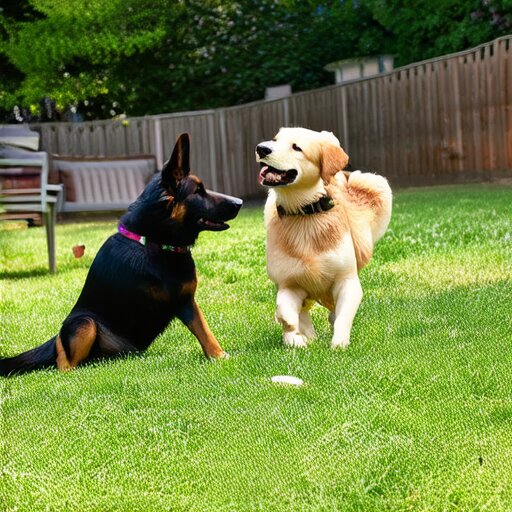}
        \end{subfigure}
        \hfill
        \begin{subfigure}[t]{0.48\textwidth}
            \centering
            \includegraphics[width=\linewidth]{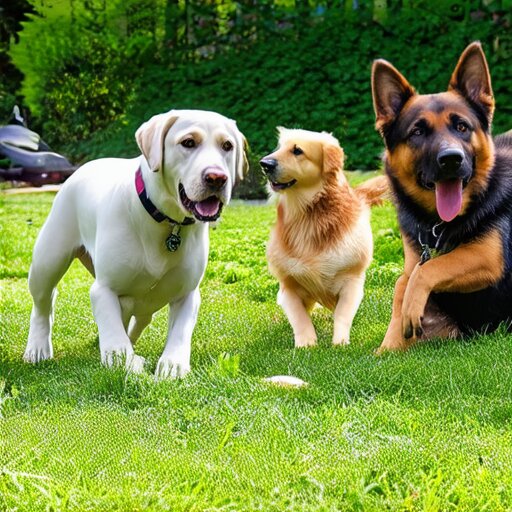}
        \end{subfigure}
    \end{subfigure}
    \caption{\textbf{Side-by-side comparison of Stable Diffusion 3.5 (left) and Stable Diffusion 3.5 + JEDI (right).} The base model often mixes attributes or omits subjects, while JEDI corrects these issues. See \cref{fig:extra_samples_sd35_1,fig:extra_samples_sd35_2} in \cref{sec:samples} for full prompts and more examples.}
    \label{fig:image_pairs}
    \vspace{-1mm}
\end{figure*}

\noindent\textbf{Objective Formulation.}
Let $S$ denote the set of subjects in the text prompt, and let $P_s$ be the set of attention maps associated with subject $s \in S$. The total loss consists of three additive components:
\begin{enumerate}
    \item \textbf{Intra-group Coherence:} Encourages attention maps within each group (e.g., between an attribute and its subject) to be similar by minimizing their JSD:
    \begin{align*}
        \frac{1}{|S|} \sum_{s \in S} \hat{D}_\mathrm{JS}(P_s).
    \end{align*}
    \item \textbf{Inter-group Separation:} For each subject $s$, we compute its mixture distribution: $\vec{m}_{s} = \frac{1}{|P_s|} \sum_{\vec{p} \in P_s} \vec{p}$. Let $M = \{\vec{m}_s \mid s \in S\}$. To encourage separation between subjects, we maximize the divergence between these mixtures, by minimizing:
    \begin{align*}
        1 - \hat{D}_\mathrm{JS}(M).
    \end{align*}
    \item \textbf{Diversity Regularization:} To avoid overly sharp or degenerate maps, we encourage spatial spread by maximizing the normalized entropy of each mixture distribution. To this end, we minimize:
    \begin{align*}
        \lambda \cdot \frac{1}{|S|} \sum_{s \in S} \lft(1 - \hat{H}(\vec{m}_s)\rgt),
    \end{align*}
    where $\lambda$ is a hyperparameter controlling the strength of the regularization term. In practice, we set $\lambda = 0.01$.
\end{enumerate}
We provide further analysis of the effect of each component in the form of an ablation study in \cref{sec:ablation}.

\begin{figure}
    \centering
    \vspace{-1mm}
    {\small \textit{``\textbf{\texttt{\char`\<}Messi\texttt{\char`\>}} and \textbf{\texttt{\char`\<}Taylor\texttt{\char`\>}} in front of Mount Fuji''}}\\
    \begin{subfigure}[t]{0.4\linewidth}
        \centering
        \includegraphics[width=\linewidth]{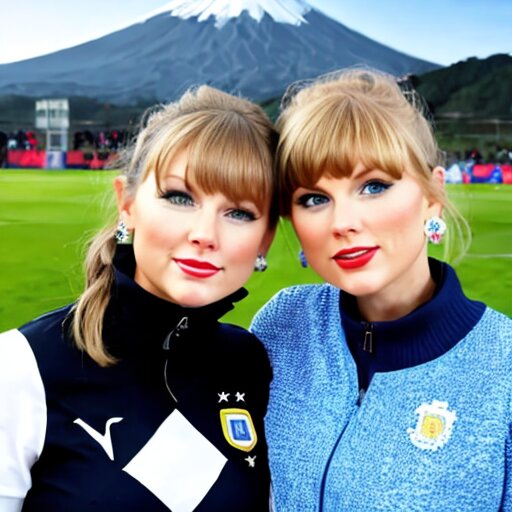}
    \end{subfigure}
    \hspace{1mm}
    \begin{subfigure}[t]{0.4\linewidth}
        \centering
        \includegraphics[width=\linewidth]{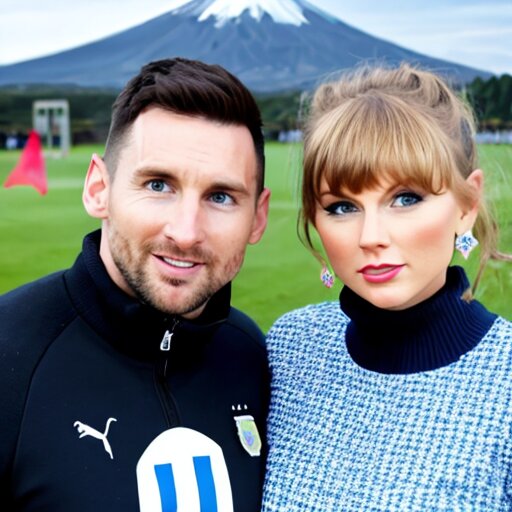}
    \end{subfigure}
    \\
    \vspace{-1mm}
    \caption{\textbf{Comparison between LoRACLR (left) and LoRACLR + JEDI (right).} The base model without a control shows attribute mixing, while JEDI produces clearer subject separation.}
    \label{fig:loraclr}\vspace{-4mm}
\end{figure}

\noindent\textbf{Update Formulation.}
To efficiently update the latent representation, we follow the Fast Gradient Sign Method \citep{goodfellow_explaining_2015} and perform:
\begin{align*}
    \vec{x}_t \leftarrow \vec{x}_t - \alpha \cdot \mathrm{sign}\left( \nabla_{\vec{x}_t} \mathrm{score}(A_t) \right),
\end{align*}
where $\mathrm{sign}(\cdot)$ is applied element-wise. This formulation accelerates updates while enabling finer control over the latent shift. We analyze the effect of $\alpha$ in \cref{fig:lr}; unless otherwise stated, we use $\alpha = 3 \times 10^{-3}$ throughout.
\section{Experiments}\label{sec:experiments}
We evaluate JEDI in three settings: 1.) a synthetic comparison against the NT-Xent loss used in the latent optimization technique CONFORM \citep{meral_conform_2023}; 2.) qualitative results on Stable Diffusion 3.5 (SD3.5)\footnote{\href{https://huggingface.co/stabilityai/stable-diffusion-3.5-medium}{huggingface.co/stabilityai/stable-diffusion-3.5-medium}}; and 3.) quantitative experiments on Stable Diffusion 1.5 (SD1.5)\footnote{\href{https://huggingface.co/stable-diffusion-v1-5/stable-diffusion-v1-5}{huggingface.co/stable-diffusion-v1-5}}, including a comparison to CONFORM and an evaluation of JEDI applied to LoRACLR, a variant of SD1.5 \citep{simsar_loraclr_2024}, to demonstrate its broader applicability. All experiments were conducted on a NVIDIA GeForce GTX TITAN X. Implementation details are provided in \cref{sec:implementation}.

\noindent\textbf{Synthetic Comparison.}
Contrastive objectives for aligning attention maps—bringing same-subject maps closer while pushing different ones apart—were first explored in CONFORM \citep{meral_conform_2023}, which uses the NT-Xent loss \citep{oord_representation_2019,he_momentum_2020,chen_simple_2020} based on cosine similarity. While effective in embedding spaces, cosine similarity is not well-suited for optimizing probability distributions: it tends to collapse mass into narrow peaks and fails to capture broader structural relationships.

To illustrate this limitation, we construct a toy example with four overlapping 1D Gaussians: blue and green represent one subject, red and purple another. The objective is to align distributions within the same group while separating those across groups. As shown in \cref{fig:ntx_vs_jsd}, JEDI preserves the support of each group, producing coherent mixtures, while NT-Xent distorts shapes and leads to over-concentrated and fragmented modes.

Since attention maps are soft spatial probability fields, preserving their continuity and avoiding artificial multimodality is critical, for example, to prevent the same subject from being generated in multiple places. JEDI’s objective maintains this structure, encouraging stable and semantically grounded attention patterns for generation.

\noindent\textbf{Stable Diffusion 3.5.}
We apply JEDI to SD3.5 and assess image quality on a custom dataset of prompts involving visually similar object pairs (e.g., \textit{``apple''} and \textit{``pear''}). For each prompt, we generate two images: one using vanilla SD3.5 and one using SD3.5 + JEDI.

As shown in \cref{fig:image_pairs}, JEDI consistently improves over the base model by correctly rendering both subjects and reducing attribute mixing. Moreover, since we set the learning rate to $\alpha = 3 \times 10^{-3}$, the overall composition and background remain nearly unchanged (e.g., in \textit{``A street bike and a dirt bike [...]''}). For additional examples, see \cref{sec:samples}.

\noindent\textbf{Comparison to CONFORM.}
To compare directly with CONFORM, originally designed for SD1.5, we adapt their implementation by replacing the CONFORM component with our JEDI objective. CONFORM performs optimization up to the 29th timestep, applying 20 iterative latent updates at steps 0, 10, and 20—totaling 69 updates. In contrast, JEDI achieves comparable or better results with just 18 updates, making it approximately $67\%$ faster in practice.

Additionally, JEDI operates with a smaller learning rate, resulting in images that remain closer to the base model's distribution. By comparison, CONFORM begins with a much higher rate ($\alpha = 20$, tapering to $16.85$), resulting in greater stylistic drift. Visual comparisons in \cref{fig:conform} highlight JEDI’s superior subject separation and overall image quality. Additional examples are provided in \cref{sec:samples}.

\noindent\textbf{Extension to LoRACLR.}
We further test JEDI on LoRACLR \citep{simsar_loraclr_2024}, a multi-concept model known to suffer from attribute mixing. On a model combining 14 distinct concepts, JEDI significantly improves subject separation (see \cref{fig:loraclr}), highlighting its flexibility across architectures. Implementation details and additional examples are in \cref{sec:implementation,sec:samples}.

\section{Discussion and Future Work}
\noindent\textbf{Unbiased Disentanglement Score.} We find that the inter-group loss term in the JEDI objective naturally serves as an effective metric for measuring subject disentanglement during generation. For the images in \cref{fig:intro_example}, the disentangled image achieves a mean JSD of $0.40 \pm 0.15$, compared to $0.17 \pm 0.10$ for the entangled counterpart, with full progression over  time shown in \cref{fig:jsd_evol}.
Unlike CLIP-based metrics, this score is computed directly from internal attention maps, making it lightweight, model-internal, and free from external supervision or bias. This presents a promising alternative for evaluating subject separation in multi-object prompts, particularly in test-time settings.

\noindent\textbf{Efficiency and Time Complexity.} 
While JEDI is highly efficient relative to existing methods, it roughly doubles inference time compared to the base model. One potential solution is to shift its objective to the training or fine-tuning stage, as it is naturally defined over all diffusion steps and requires no supervision. This would reduce inference time while preserving the benefits of the JEDI framework. 


\bibliography{references}

\begin{thebibliography}{19}
\providecommand{\natexlab}[1]{#1}
\providecommand{\url}[1]{\texttt{#1}}
\expandafter\ifx\csname urlstyle\endcsname\relax
  \providecommand{\doi}[1]{doi: #1}\else
  \providecommand{\doi}{doi: \begingroup \urlstyle{rm}\Url}\fi

\bibitem[Binyamin et~al.(2024)Binyamin, Tewel, Segev, Hirsch, Rassin, and Chechik]{binyamin_make_2024}
Binyamin, L., Tewel, Y., Segev, H., Hirsch, E., Rassin, R., and Chechik, G.
\newblock Make it count: Text-to-image generation with an accurate number of objects.
\newblock \emph{arXiv preprint arXiv:2406.10210}, 2024.

\bibitem[Chefer et~al.(2023)Chefer, Alaluf, Vinker, Wolf, and Cohen-Or]{chefer_attend-and-excite_2023}
Chefer, H., Alaluf, Y., Vinker, Y., Wolf, L., and Cohen-Or, D.
\newblock Attend-and-excite: Attention-based semantic guidance for text-to-image diffusion models.
\newblock \emph{ACM transactions on Graphics (TOG)}, 42\penalty0 (4):\penalty0 1--10, 2023.

\bibitem[Chen et~al.(2020)Chen, Kornblith, Norouzi, and Hinton]{chen_simple_2020}
Chen, T., Kornblith, S., Norouzi, M., and Hinton, G.
\newblock A simple framework for contrastive learning of visual representations.
\newblock In \emph{International conference on machine learning}, pp.\  1597--1607. PmLR, 2020.

\bibitem[Esser et~al.(2024)Esser, Kulal, Blattmann, Entezari, M{\"u}ller, Saini, Levi, Lorenz, Sauer, Boesel, et~al.]{esser_scaling_2024}
Esser, P., Kulal, S., Blattmann, A., Entezari, R., M{\"u}ller, J., Saini, H., Levi, Y., Lorenz, D., Sauer, A., Boesel, F., et~al.
\newblock Scaling rectified flow transformers for high-resolution image synthesis.
\newblock In \emph{Forty-first international conference on machine learning}, 2024.

\bibitem[Goodfellow et~al.(2014)Goodfellow, Shlens, and Szegedy]{goodfellow_explaining_2015}
Goodfellow, I.~J., Shlens, J., and Szegedy, C.
\newblock Explaining and harnessing adversarial examples.
\newblock \emph{arXiv preprint arXiv:1412.6572}, 2014.

\bibitem[Gu et~al.(2023)Gu, Wang, Wu, Shi, Chen, Fan, Xiao, Zhao, Chang, Wu, et~al.]{gu_mix--show_2023}
Gu, Y., Wang, X., Wu, J.~Z., Shi, Y., Chen, Y., Fan, Z., Xiao, W., Zhao, R., Chang, S., Wu, W., et~al.
\newblock Mix-of-show: Decentralized low-rank adaptation for multi-concept customization of diffusion models.
\newblock \emph{Advances in Neural Information Processing Systems}, 36:\penalty0 15890--15902, 2023.

\bibitem[He et~al.(2020)He, Fan, Wu, Xie, and Girshick]{he_momentum_2020}
He, K., Fan, H., Wu, Y., Xie, S., and Girshick, R.
\newblock Momentum contrast for unsupervised visual representation learning.
\newblock In \emph{Proceedings of the IEEE/CVF conference on computer vision and pattern recognition}, pp.\  9729--9738, 2020.

\bibitem[Ho et~al.(2020)Ho, Jain, and Abbeel]{ho_denoising_2020}
Ho, J., Jain, A., and Abbeel, P.
\newblock Denoising diffusion probabilistic models.
\newblock \emph{Advances in neural information processing systems}, 33:\penalty0 6840--6851, 2020.

\bibitem[Kwon et~al.(2024)Kwon, Jenni, Li, Lee, Ye, and Heilbron]{kwon_concept_2024}
Kwon, G., Jenni, S., Li, D., Lee, J.-Y., Ye, J.~C., and Heilbron, F.~C.
\newblock Concept weaver: Enabling multi-concept fusion in text-to-image models.
\newblock In \emph{Proceedings of the IEEE/CVF Conference on Computer Vision and Pattern Recognition}, pp.\  8880--8889, 2024.

\bibitem[Lipman et~al.(2022)Lipman, Chen, Ben-Hamu, Nickel, and Le]{lipman2022flow}
Lipman, Y., Chen, R.~T., Ben-Hamu, H., Nickel, M., and Le, M.
\newblock Flow matching for generative modeling.
\newblock \emph{arXiv preprint arXiv:2210.02747}, 2022.

\bibitem[Liu et~al.(2022)Liu, Li, Du, Torralba, and Tenenbaum]{liu_compositional_2023}
Liu, N., Li, S., Du, Y., Torralba, A., and Tenenbaum, J.~B.
\newblock Compositional visual generation with composable diffusion models.
\newblock In \emph{European Conference on Computer Vision}, pp.\  423--439. Springer, 2022.

\bibitem[Meral et~al.(2024)Meral, Simsar, Tombari, and Yanardag]{meral_conform_2023}
Meral, T. H.~S., Simsar, E., Tombari, F., and Yanardag, P.
\newblock Conform: Contrast is all you need for high-fidelity text-to-image diffusion models.
\newblock In \emph{Proceedings of the IEEE/CVF Conference on Computer Vision and Pattern Recognition}, pp.\  9005--9014, 2024.

\bibitem[Oord et~al.(2018)Oord, Li, and Vinyals]{oord_representation_2019}
Oord, A. v.~d., Li, Y., and Vinyals, O.
\newblock Representation learning with contrastive predictive coding.
\newblock \emph{arXiv preprint arXiv:1807.03748}, 2018.

\bibitem[Peebles \& Xie(2023)Peebles and Xie]{peebles_scalable_2023}
Peebles, W. and Xie, S.
\newblock Scalable diffusion models with transformers.
\newblock In \emph{Proceedings of the IEEE/CVF international conference on computer vision}, pp.\  4195--4205, 2023.

\bibitem[Podell et~al.(2023)Podell, English, Lacey, Blattmann, Dockhorn, M{\"u}ller, Penna, and Rombach]{podell_sdxl_2023}
Podell, D., English, Z., Lacey, K., Blattmann, A., Dockhorn, T., M{\"u}ller, J., Penna, J., and Rombach, R.
\newblock Sdxl: Improving latent diffusion models for high-resolution image synthesis.
\newblock \emph{arXiv preprint arXiv:2307.01952}, 2023.

\bibitem[Ramesh et~al.(2022)Ramesh, Dhariwal, Nichol, Chu, and Chen]{ramesh_hierarchical_2022}
Ramesh, A., Dhariwal, P., Nichol, A., Chu, C., and Chen, M.
\newblock Hierarchical text-conditional image generation with clip latents.
\newblock \emph{arXiv preprint arXiv:2204.06125}, 1\penalty0 (2):\penalty0 3, 2022.

\bibitem[Rombach et~al.(2022)Rombach, Blattmann, Lorenz, Esser, and Ommer]{rombach_high-resolution_2022}
Rombach, R., Blattmann, A., Lorenz, D., Esser, P., and Ommer, B.
\newblock High-resolution image synthesis with latent diffusion models.
\newblock In \emph{Proceedings of the IEEE/CVF conference on computer vision and pattern recognition}, pp.\  10684--10695, 2022.

\bibitem[Simsar et~al.(2024)Simsar, Hofmann, Tombari, and Yanardag]{simsar_loraclr_2024}
Simsar, E., Hofmann, T., Tombari, F., and Yanardag, P.
\newblock Loraclr: Contrastive adaptation for customization of diffusion models.
\newblock \emph{arXiv preprint arXiv:2412.09622}, 2024.

\bibitem[Wei et~al.(2024)Wei, Chen, Zhou, and Pan]{wei_enhancing_2024}
Wei, T., Chen, D., Zhou, Y., and Pan, X.
\newblock Enhancing mmdit-based text-to-image models for similar subject generation.
\newblock \emph{arXiv preprint arXiv:2411.18301}, 2024.

\end{thebibliography}
\bibliographystyle{icml2025}

\newpage
\appendix
\section{Hyperparameter}
The learning rate $\alpha$ serves as a critical hyperparameter in our optimization, especially due to the use of the $\mathrm{sign}(\cdot)$ function, which restricts the gradient to unit magnitude.

\begin{figure*}[t]
    \centering
    \begin{subfigure}[b]{0.19\textwidth}
        \centering
        \captionsetup{labelformat=empty}
        \caption{$\bm{\alpha = 5 \times 10^{-1}}$}
        \includegraphics[width=\textwidth]{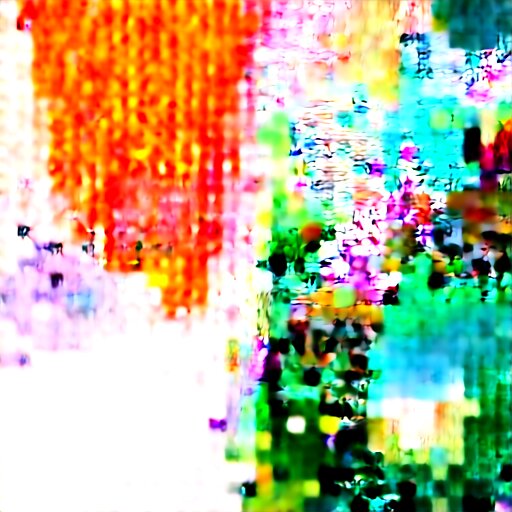}
    \end{subfigure}\hfill
    \begin{subfigure}[b]{0.19\textwidth}
        \centering
        \captionsetup{labelformat=empty}
        \caption{$\alpha = 3 \times 10^{-1}$}
        \includegraphics[width=\textwidth]{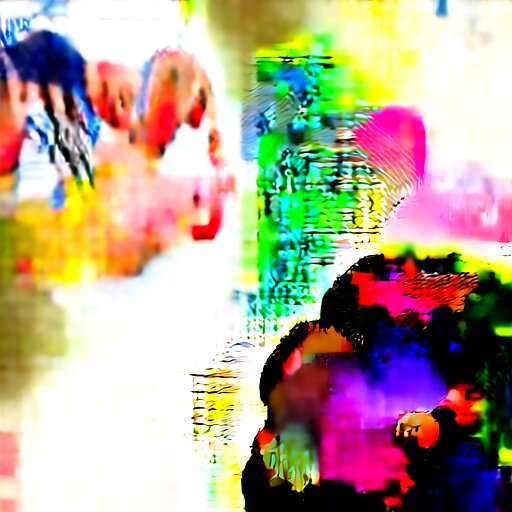}
    \end{subfigure}\hfill
    \begin{subfigure}[b]{0.19\textwidth}
        \centering
        \captionsetup{labelformat=empty}
        \caption{$\alpha = 1 \times 10^{-1}$}
        \includegraphics[width=\textwidth]{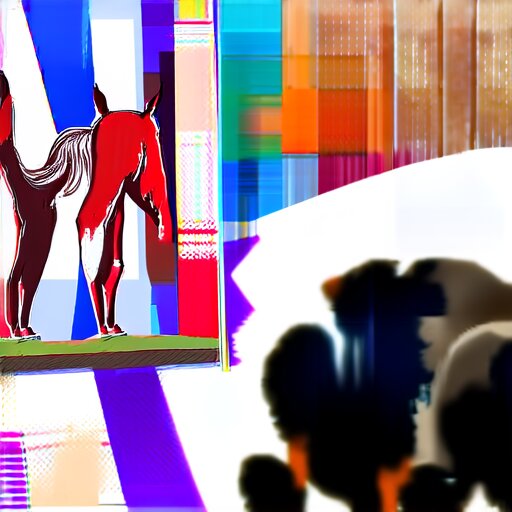}
    \end{subfigure}\hfill
    \begin{subfigure}[b]{0.19\textwidth}
        \centering
        \captionsetup{labelformat=empty}
        \caption{$\alpha = 5 \times 10^{-2}$}
        \includegraphics[width=\textwidth]{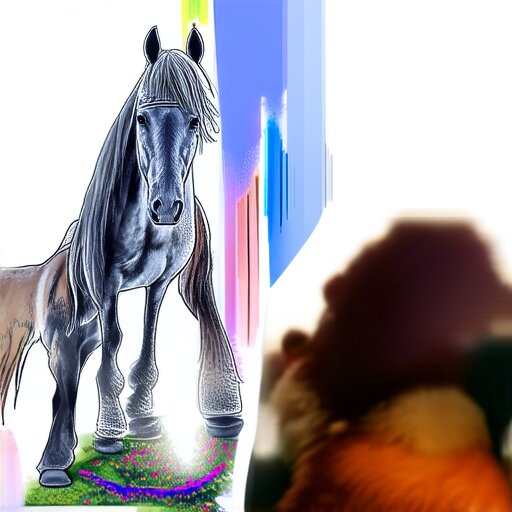}
    \end{subfigure}\hfill
    \begin{subfigure}[b]{0.19\textwidth}
        \centering
        \captionsetup{labelformat=empty}
        \caption{$\alpha = 3 \times 10^{-2}$}
        \includegraphics[width=\textwidth]{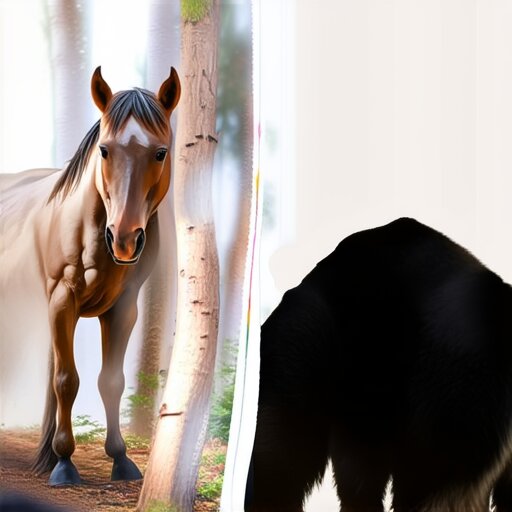}
    \end{subfigure} \\
    \vspace{5pt}
    \begin{subfigure}[b]{0.19\textwidth}
        \centering
        \captionsetup{labelformat=empty}
        \includegraphics[width=\textwidth]{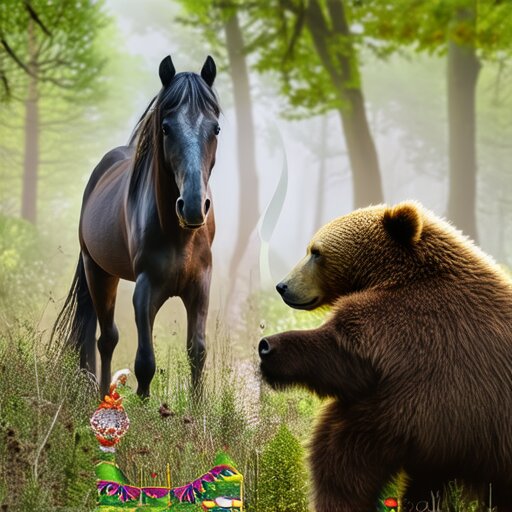}
        \caption{$\alpha = 1 \times 10^{-2}$}
    \end{subfigure}\hfill
    \begin{subfigure}[b]{0.19\textwidth}
        \centering
        \captionsetup{labelformat=empty}
        \includegraphics[width=\textwidth]{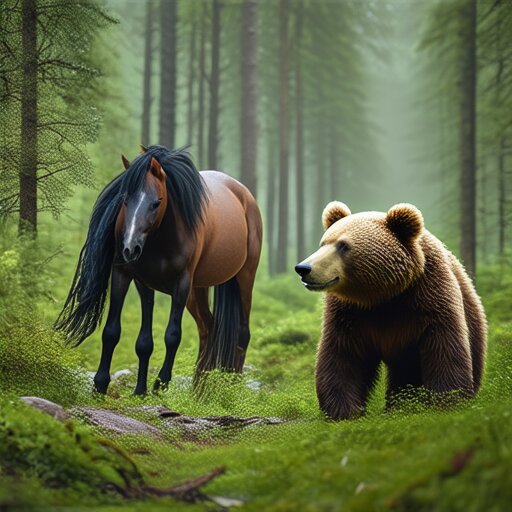}
        \caption{$\alpha = 5 \times 10^{-3}$}
    \end{subfigure}\hfill
    \begin{subfigure}[b]{0.19\textwidth}
        \centering
        \captionsetup{labelformat=empty}
        \includegraphics[width=\textwidth]{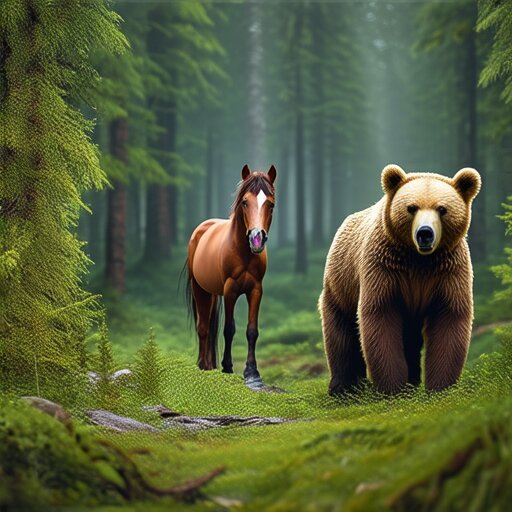}
        \caption{$\alpha = 3 \times 10^{-3}$}
    \end{subfigure}\hfill
    \begin{subfigure}[b]{0.19\textwidth}
        \centering
        \captionsetup{labelformat=empty}
        \includegraphics[width=\textwidth]{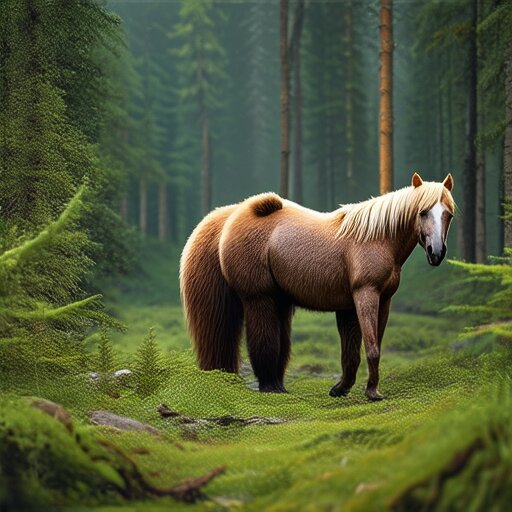}
        \caption{$\alpha = 1 \times 10^{-3}$}
    \end{subfigure}\hfill
    \begin{subfigure}[b]{0.19\textwidth}
        \centering
        \captionsetup{labelformat=empty}
        \includegraphics[width=\textwidth]{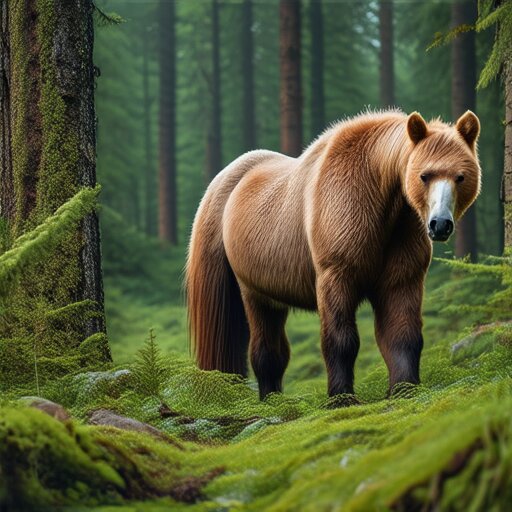}
        \caption{Base Model}
    \end{subfigure}\hfill
    \caption{\textbf{Effect of learning rate $\alpha$ on image generation.} Outputs generated for the same prompt, ``\textit{A horse and a bear in a forest},'' using Stable Diffusion 3.5 under identical settings, varying only the learning rate. Higher $\alpha$ values lead to excessive changes in the latent space, deteriorating image quality, while lower values result in minimal updates and limited visual difference.}
    \label{fig:lr}
\end{figure*}

As such, $\alpha$ directly controls the extent to which the latent image is updated at each step. To illustrate this effect, we show in \cref{fig:lr} the outputs of JEDI applied to Stable Diffusion 3.5 under varying values of $\alpha$. As expected, large learning rates lead to overly aggressive updates, causing the optimization to diverge and fail to produce coherent images. Conversely, excessively small values have a negligible effect, resulting in little to no noticeable changes.

\section{Proofs}\label{sec:proofs}

\begin{lemma}[Upper Bound of Jensen-Shannon Divergence]\label{lem:jsd}
    Let $P = \{\vec{p}^{(1)}, \dots, \vec{p}^{(n)}\} \subset \R^d$ be a set of probability distributions. Then, $D_\mathrm{JS}(P)$ is upper bounded by $\log n$.   
\end{lemma}
\begin{proof}
    Define $P$ as in \cref{lem:jsd}, then the JSD is defined as follows:
    \begin{align*}
        D_\mathrm{JS}(P) = \frac{1}{n}\sum_{k=1}^{n} D_\mathrm{KL}\lft(\vec{p}^{(k)} \;\|\; \vec{m}\rgt), \quad \vec{m} = \frac{1}{n}\sum_{k=1}^{n} \vec{p}^{(k)}.
    \end{align*}
    We can upper bound each $D_\mathrm{KL}$-term as follows:
    \begin{align*}
        D_\mathrm{KL} (\vec{p}^{(k)} \;\|\; \vec{m}) &= \sum_{i=1}^d p^{(k)}_i \log \frac{p^{(k)}_i}{m_i}\\
        &= \sum_{i=1}^d {p}^{(k)}_i \log \frac{{p}^{(k)}_i}{\frac{1}{n}\sum_{\ell=1}^n p^{(\ell)}_i}\\
        &= \sum_{i=1}^d {p}^{(k)}_i \log \lft( n \cdot \frac{p^{(k)}_i}{\sum_{\ell=1}^n p^{(\ell)}_i} \rgt)\\
        &\leq \sum_{i=1}^d {p}^{(k)}_i \log n\\
        &= \log n.
    \end{align*}
    Plugging this bound back into the definition of the JSD, yields the desired results:
    \begin{align*}
         \frac{1}{n}\sum_{k=1}^{n} D_\mathrm{KL}\lft(\vec{p}^{(k)} \;\|\; \vec{m}\rgt) \leq \frac{1}{n}\sum_{k=1}^{n} \log n = \log n
    \end{align*}
\end{proof}

\begin{lemma}[Upper Bound of Shannon Entropy]\label{lem:entropy}
    Let $\vec{p} \in \R^d$ be a discrete probability distribution, such that $p_i \geq 0$ and $\sum_i^d p_i = 1$. Then, its entropy $H(\vec{p})$ is upper bounded by $\log d$.
\end{lemma}

\begin{proof}
    Let $\vec{p}$ be defined as in \cref{lem:entropy}. We define a Lagrangian as follows:
    \begin{align*}
        \mathcal{L}(\vec{p}, \lambda) = H(\vec{p}) + \lambda \cdot \lft(1 - \sum_i^d p_i \rgt),
    \end{align*}
    where $\lambda \in \R$ is a Langrage multiplier.
    Taking the derivative with respect to each $p_i$ and setting it to zero yields:
    \begin{align*}
        \nabla_{p_i} \mathcal{L}(\vec{p}, \lambda) = 0 \iff \log(p_i) = \lambda - 1.
    \end{align*}
    Thus, all $p_i$ must be equal at the maximum. Using the constraint $\sum_{i=1}^d p_i = 1$, it follows that $p_i = \frac{1}{d}$ for all $i$.  
    
    Substituting this result back into the definition of entropy gives:
    \begin{align*}
        H(\vec{p}) = -\sum_{i=1}^d p_i \log(p_i) = - \sum_{i=1}^d \frac{1}{d} \log(\frac{1}{d}) = \log(d).
    \end{align*}
\end{proof}

\section{Pseudo-code of JEDI}\label{sec:algo}
\Cref{alg:jedi} illustrates a minimal implementation of JEDI's test-time adaptation procedure, integrated into a standard iterative denoising loop of a diffusion model. The modifications introduced by JEDI are highlighted in blue, while the rest of the loop corresponds to the denoising process.

At each timestep, the model produces a denoised latent $\vec{x}_{t+1}$ along with the corresponding internal attention maps $A_t$.

\begin{algorithm}[H]
\caption{JEDI Test-time Adaptation}
\begin{algorithmic}[1]
    \STATE \textbf{Input:} Condition prompt $\vec{c}$
    \STATE $\vec{x}_0 \sim \mathcal{N}\lft(\mat{0}, \mat{I}\rgt)$
    \FOR{$t = 0$ to $T-1$}
        \color{blue}\IF{$t \leq K$}
            \STATE \underline{\hspace{3mm}}, $A_t \gets \texttt{Model}(\vec{x}_{t}, \vec{c})$
            \STATE $\vec{x}_{t} \gets \vec{x}_{t} - \alpha \cdot \texttt{sign}(\nabla_{\vec{x}_{t}}\texttt{JEDI}(A_t, \vec{c}))$
        \ENDIF
        \color{black}\STATE $\vec{x}_{t+1}, \underline{\hspace{3mm}} \gets \texttt{Model}(\vec{x}_{t}, \vec{c})$
    \ENDFOR
    \STATE \textbf{return} $\vec{x}_T$
\end{algorithmic}
\label{alg:jedi}
\end{algorithm}

\section{Ablation}\label{sec:ablation}
\begin{figure*}[t]
    \centering
    \begin{subfigure}[b]{0.19\textwidth}
        \centering
        \captionsetup{labelformat=empty}
        \caption{JEDI}
        \includegraphics[width=\textwidth]{figures/pairs/bear_horse/jsd/image_0.jpg}
    \end{subfigure}\hfill
    \begin{subfigure}[b]{0.19\textwidth}
        \centering
        \captionsetup{labelformat=empty,justification=centering}
        \caption{Without\\Intra-group Coherence}
        \includegraphics[width=\textwidth]{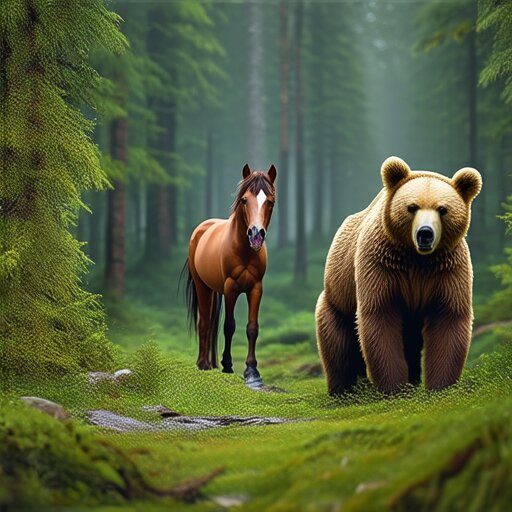}
    \end{subfigure}\hfill
    \begin{subfigure}[b]{0.19\textwidth}
        \centering
        \captionsetup{labelformat=empty,justification=centering}
        \caption{Without\\Inter-group Separation}
        \includegraphics[width=\textwidth]{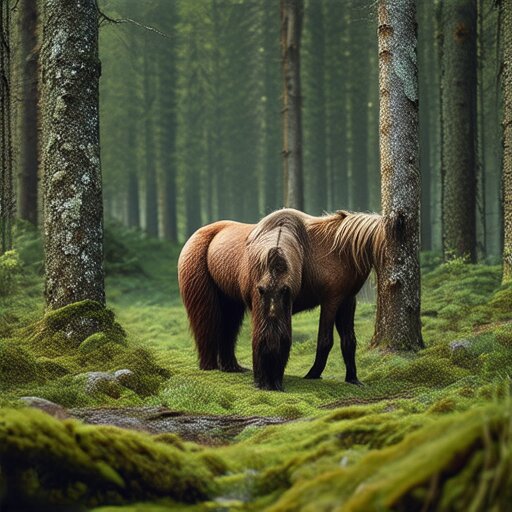}
    \end{subfigure}\hfill
    \begin{subfigure}[b]{0.19\textwidth}
        \centering
        \captionsetup{labelformat=empty,justification=centering}
        \caption{Without\\Diversity Reguralization}
        \includegraphics[width=\textwidth]{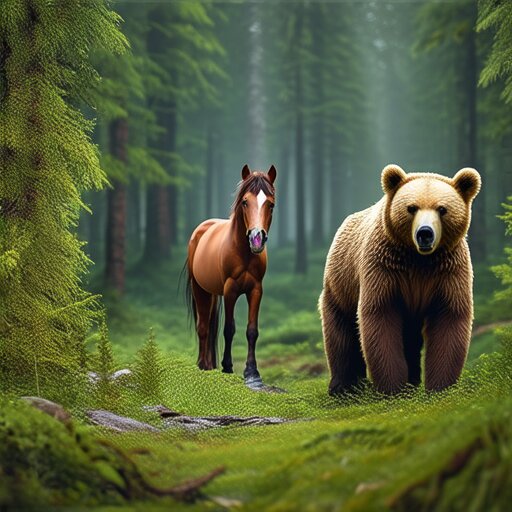}
    \end{subfigure}\hfill
    \begin{subfigure}[b]{0.19\textwidth}
        \centering
        \captionsetup{labelformat=empty}
        \caption{Base}
        \includegraphics[width=\textwidth]{figures/pairs/bear_horse/base/image_0.jpg}
    \end{subfigure}\hfill
    \par
    \vspace{-0.5mm}\textit{``A \textbf{horse} and a \textbf{bear} in a forest''}\par\vspace{1.5mm}
    \begin{subfigure}[b]{0.19\textwidth}
        \centering
        \includegraphics[width=\textwidth]{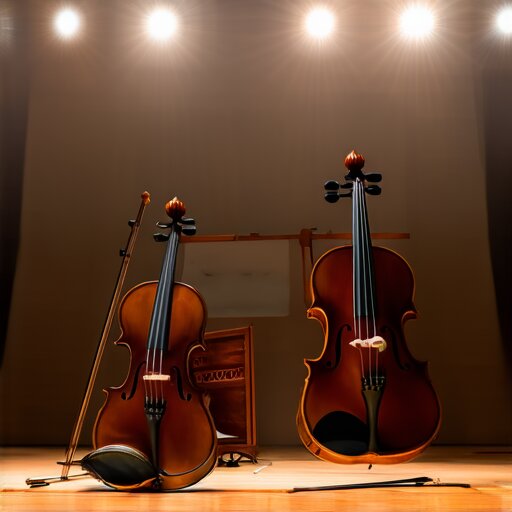}
    \end{subfigure}\hfill
    \begin{subfigure}[b]{0.19\textwidth}
        \centering
        \captionsetup{labelformat=empty}
        \includegraphics[width=\textwidth]{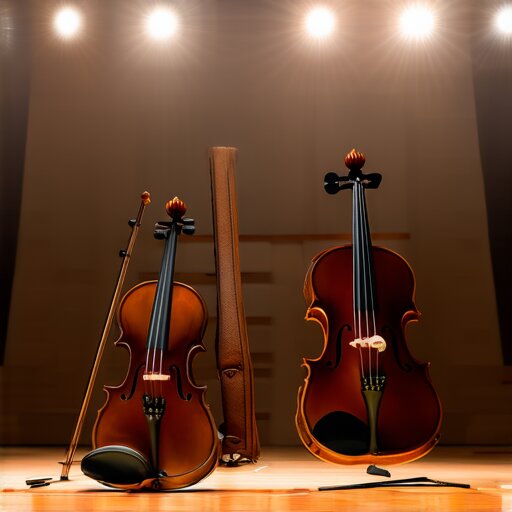}
    \end{subfigure}\hfill
    \begin{subfigure}[b]{0.19\textwidth}
        \centering
        \includegraphics[width=\textwidth]{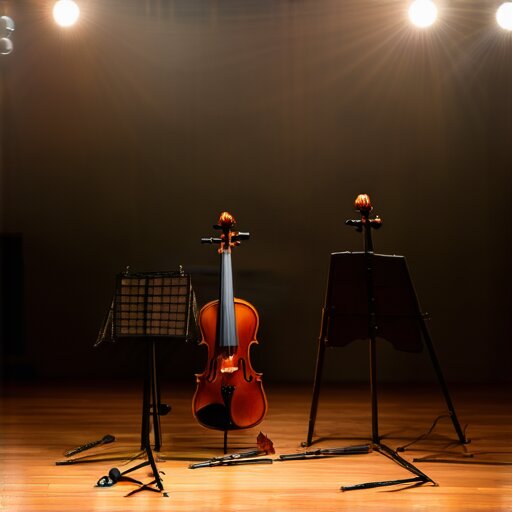}
    \end{subfigure}\hfill
    \begin{subfigure}[b]{0.19\textwidth}
        \centering
        \captionsetup{labelformat=empty}
        \includegraphics[width=\textwidth]{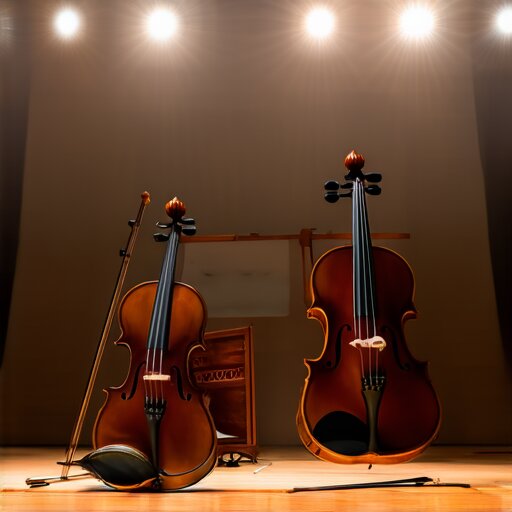}
    \end{subfigure}\hfill
    \begin{subfigure}[b]{0.19\textwidth}
        \centering
        \includegraphics[width=\textwidth]{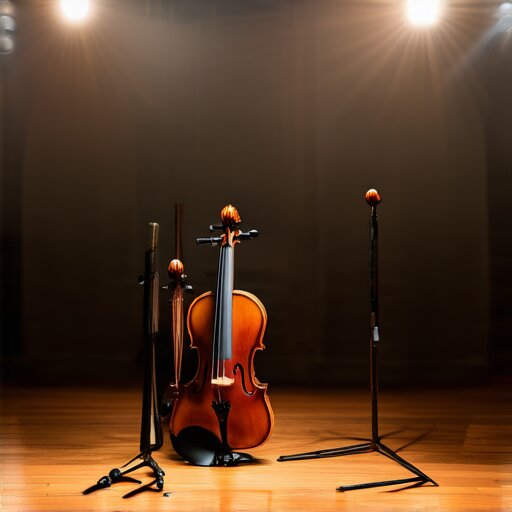}
    \end{subfigure}\hfill
    \par
    \vspace{-0.5mm}\textit{``A \textbf{violin} and a \textbf{viola} on a wooden stage under soft spotlights''}\par\vspace{1.5mm}
    \begin{subfigure}[b]{0.19\textwidth}
        \centering
        \includegraphics[width=\textwidth]{figures/triples/jedi/image_27.jpg}
    \end{subfigure}\hfill
    \begin{subfigure}[b]{0.19\textwidth}
        \centering
        \captionsetup{labelformat=empty}
        \includegraphics[width=\textwidth]{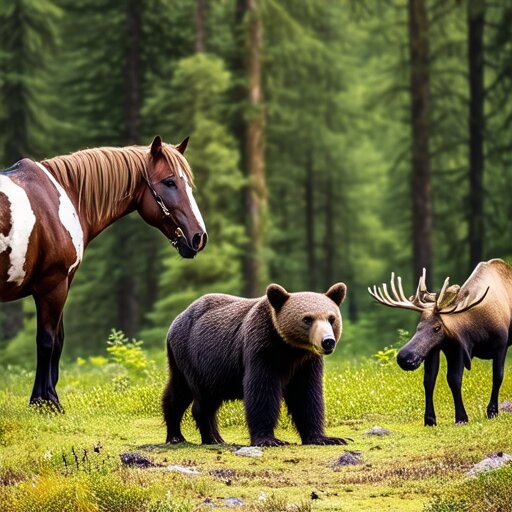}
    \end{subfigure}\hfill
    \begin{subfigure}[b]{0.19\textwidth}
        \centering
        \includegraphics[width=\textwidth]{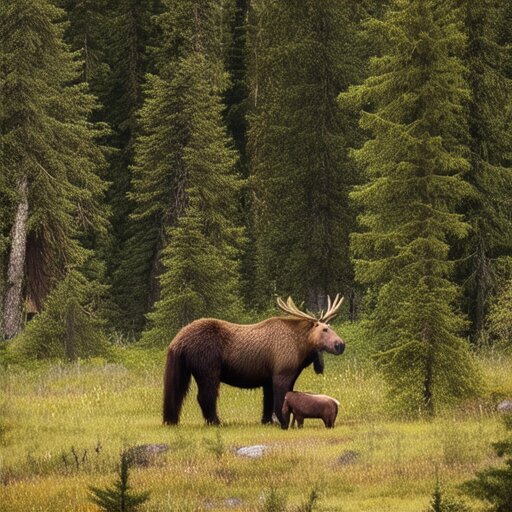}
    \end{subfigure}\hfill
    \begin{subfigure}[b]{0.19\textwidth}
        \centering
        \captionsetup{labelformat=empty}
        \includegraphics[width=\textwidth]{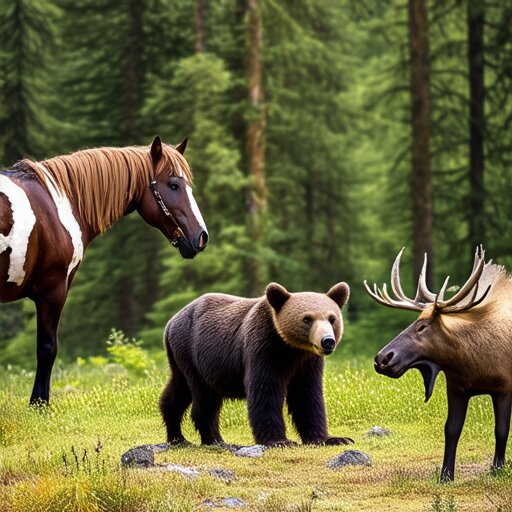}
    \end{subfigure}\hfill
    \begin{subfigure}[b]{0.19\textwidth}
        \centering
        \includegraphics[width=\textwidth]{figures/triples/base/image_27.jpg}
    \end{subfigure}\hfill
    \vspace{-0.5mm}\par\textit{``A \textbf{horse}, a \textbf{bear}, and a \textbf{moose} in a forest clearing''}
    \par\vspace{1.5mm}
    \caption{\textbf{Effect of individual components in the JEDI objective.} Outputs generated from the same prompt using Stable Diffusion 3.5 under identical sampling settings. The left column shows results with all components enabled. The middle columns each omit one component of the JEDI objective. The right column shows outputs from the base model without any JEDI adaptation.}
    \label{fig:ablation}
\end{figure*}

The JEDI objective comprises three additive components: \textit{Intra-group Coherence}, \textit{Inter-group Separation}, and a \textit{Diversity Regularizer}. To evaluate the individual contribution of each term, we conduct an ablation study by systematically removing one component at a time. The results are shown in \cref{fig:ablation}.

Overall, the best results are achieved when all three components are included. Among them, \textit{Inter-group Separation} has the most pronounced effect. This term encourages the model to spatially disentangle subjects, thereby reducing attribute mixing. Whenever it is removed, we observe noticeable shifts in image style and a significant increase in attribute mixing and spatial overlap between entities.

The effect of \textit{Intra-group Coherence} is more subtle but still important. For example, in the generation of the \textit{``moose''} subject, removing this term results in unnatural proportions. We attribute this degradation to misalignment between attention distributions produced by Stable Diffusion 3.5’s dual text encoders (T5 and CLIP) which differ substantially in architecture and semantic representation. The coherence term helps align these internal representations, yielding more consistent subject rendering.

Finally, the contribution of the \textit{Diversity Regularizer} is minimal in this setting. We scale this term with a small coefficient of $\lambda = 1 \times 10^{-2}$, which limits its influence during optimization. However, we found it to be beneficial in synthetic scenarios where we noticed attention map collapse. For this reason, we retain it as a safeguard.

\section{Implementation Details}\label{sec:implementation}
To facilitate reproducibility, we describe the key implementation details for each architecture evaluated. Full source code and experimental configurations are available on our project website: \href{https://ericbill21.github.io/JEDI/}{ericbill21.github.io/JEDI/}.

\noindent\textbf{Stable Diffusion 1.5.}
To enable a direct comparison with CONFORM \citep{meral_conform_2023}, we adopt their implementation and replace the CONFORM module with our JEDI objective. Following their experimental setup, we sample the model for $50$ timesteps using a guidance scale of $7.5$.

During each forward pass, we extract cross-attention maps at a resolution of $16 \times 16$ and compute the JEDI objective over these maps. We then backpropagate the loss and update the latent variables using signed gradients, with a learning rate of $\alpha = 3 \times 10^{-3}$. Optimization is applied only during the first 18 timesteps, which already yields strong results. We do not perform extensive hyperparameter tuning, as our primary focus is on evaluating JEDI with SD3.5.

For LoRACLR \citep{simsar_loraclr_2024}, which is built on the Mix-of-Show codebase \citep{gu_mix--show_2023}, we extend the CONFORM implementation to operate within this framework by applying the same setup used for SD 1.5. The only modification is an extended optimization window of 30 timesteps to account for the observed increased attribute mixing in LoRACLR.

\noindent\textbf{Stable Diffusion 3.5.} Modern T2I models like Stable Diffusion 3.5 are based on the Diffusion Transformer (DiT) architecture by \citet{peebles_scalable_2023}, which replaces the traditional U-Net with a sequence of DiT blocks. Unlike U-Nets, DiT does not use explicit cross-attention between image and text tokens, making it more challenging to extract spatial attention distributions for individual prompt tokens. To approximate this behavior, we took inspiration from \citet{wei_enhancing_2024}.

Each DiT block processes image tokens $\mat{X} \in \R^{n \times d}$ and prompt tokens $\mat{C} \in \R^{m \times d}$ separately, producing respective query, key, and value matrices:
\begin{align*}
    \mat{Q}_x, \mat{K}_x, \mat{V}_x \text{ (image)}\quad\text{and}\quad \mat{Q}_c, \mat{K}_c, \mat{V}_c \text{ (text)}.
\end{align*}
These matrices are then concatenated to form the full attention inputs:
\begin{align*}
    \mat{Q} &= \mathrm{concat}[\mat{Q}_x, \mat{Q}_c], \\
    \mat{K} &= \mathrm{concat}[\mat{K}_x, \mat{K}_c], \\
    \mat{V} &= \mathrm{concat}[\mat{V}_x, \mat{V}_c].   
\end{align*}
Self-attention is applied over the combined sequence, yielding the attention matrix $\mat{A} = \mathrm{softmax}(\mat{Q} \mat{K}^\top) \in \R^{(n+m) \times (n+m)}$. To estimate the spatial influence of prompt token $i$ on the image, we compute:
\begin{align*}
    \frac{1}{\sqrt{2}} \lft( \mat{A}_{n+i, :n} + \transpose{\mat{A}_{:n, n+i}} \rgt).
\end{align*}
Since this expression is not guaranteed to form a normalized distribution, we consider two options: 1.) renormalize the result, or 2.) bypass the softmax during attention and apply it only during extraction, using raw logits. Empirically, we find the second approach yields more stable and consistent results.

SD3.5 contains 24 DiT blocks, each producing an attention map per prompt token. However, not all blocks provide equally useful information. Based on visual analysis and computational efficiency, we select blocks 5 to 15 for both extraction and optimization. See \cref{fig:attn_bear,fig:attn_horse} for example visualizations.

We find that applying latent optimization during the first 18 timesteps is sufficient. All experiments use these settings, along with 28 inference steps in total and a guidance scale of 4.5, following the official recommendations.

\newcommand{\attntopbear}[1]{
    \begin{subfigure}[t]{0.07\linewidth}
        \captionsetup{labelformat=empty}
        \caption{\the\numexpr #1 + 1\relax.}%
        \includegraphics[width=\linewidth]{figures/attn_viz/attn_bear#1.jpg}%
    \end{subfigure}
}

\newcommand{\attnbottombear}[1]{
    \begin{subfigure}[t]{0.07\linewidth}
        \captionsetup{labelformat=empty}
        \includegraphics[width=\linewidth]{figures/attn_viz/attn_bear#1.jpg}%
        \caption{\the\numexpr #1 + 1\relax.}%
    \end{subfigure}
}

\begin{figure*}
    \makebox[\linewidth][c]{%
        \attntopbear{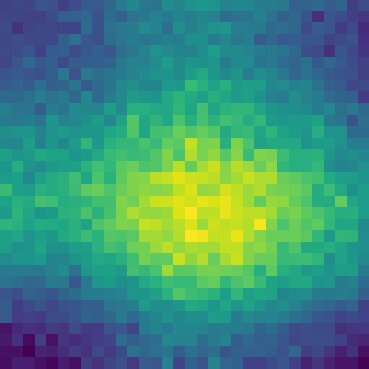}%
        \attntopbear{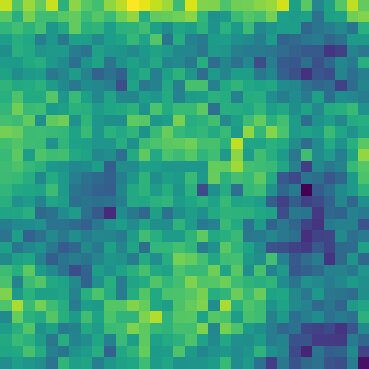}%
        \attntopbear{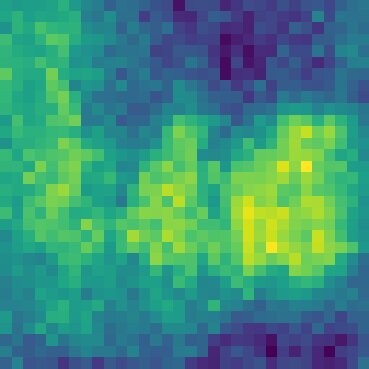}%
        \attntopbear{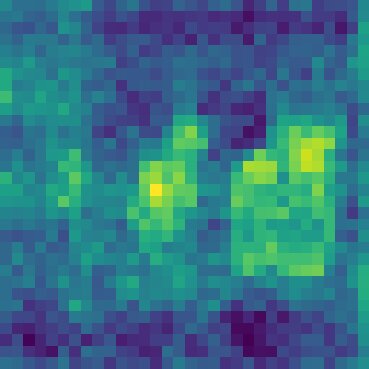}%
        \attntopbear{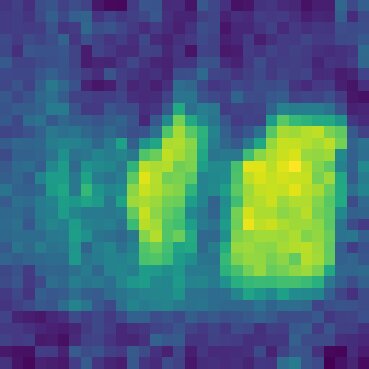}%
        \attntopbear{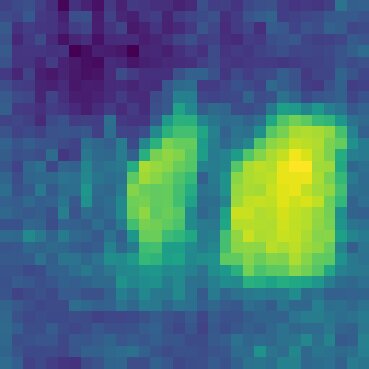}%
        \attntopbear{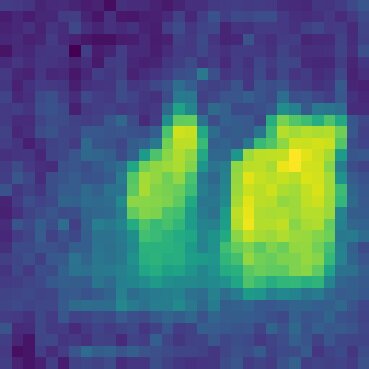}%
        \attntopbear{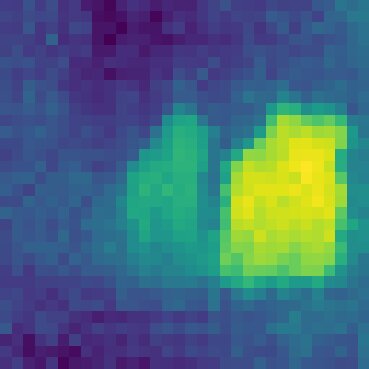}%
        \attntopbear{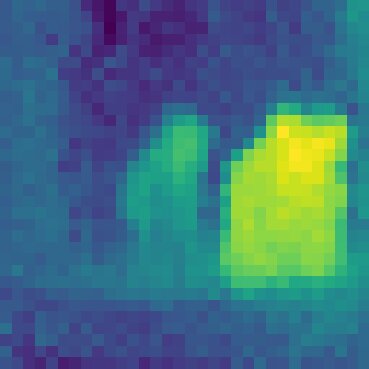}%
        \attntopbear{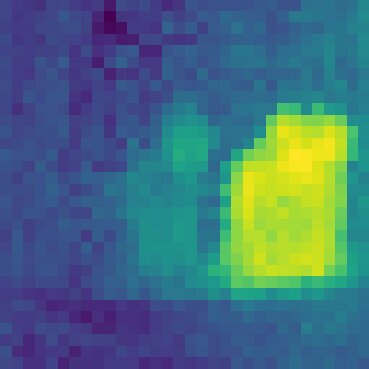}%
        \attntopbear{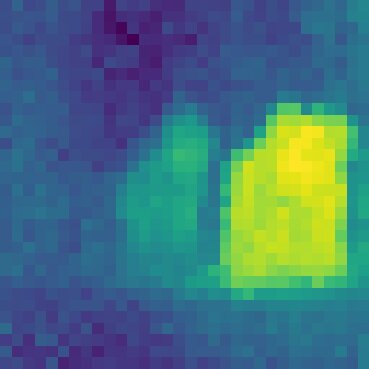}%
        \attntopbear{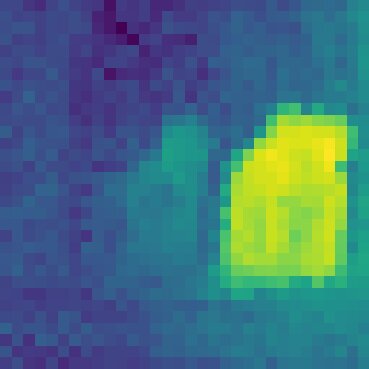}%
    }
    \makebox[\linewidth][c]{%
        \attnbottombear{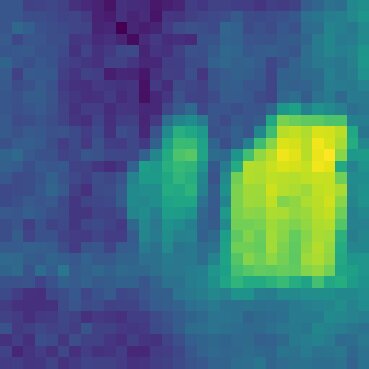}%
        \attnbottombear{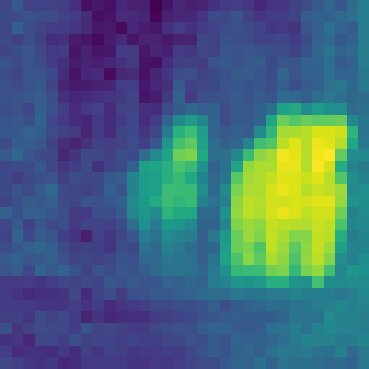}%
        \attnbottombear{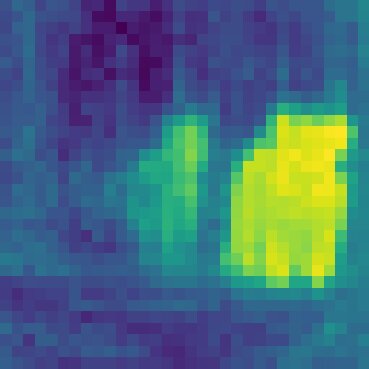}%
        \attnbottombear{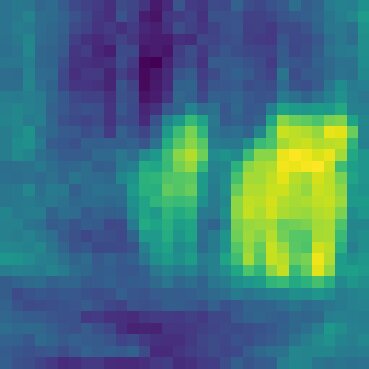}%
        \attnbottombear{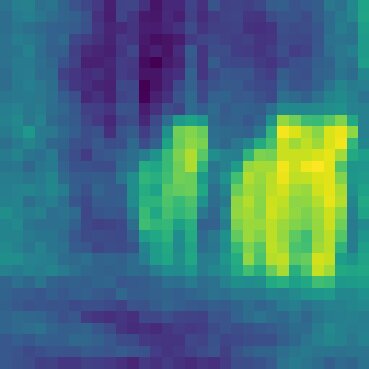}%
        \attnbottombear{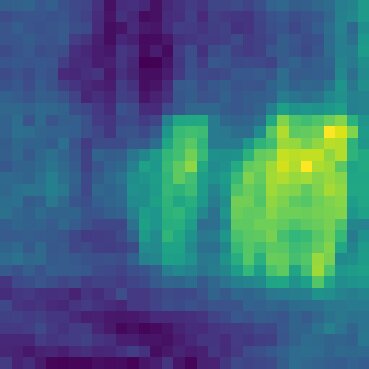}%
        \attnbottombear{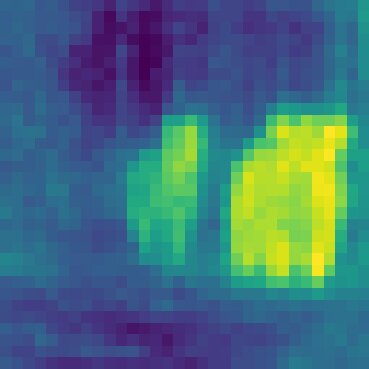}%
        \attnbottombear{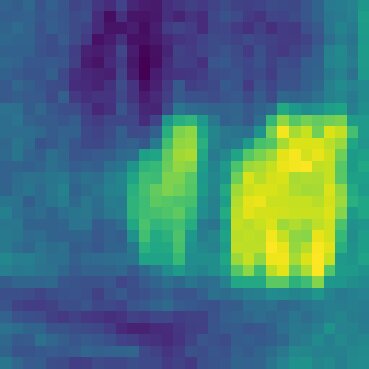}%
        \attnbottombear{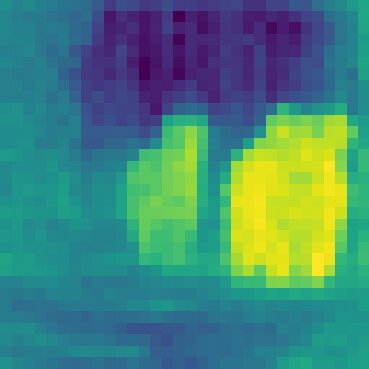}%
        \attnbottombear{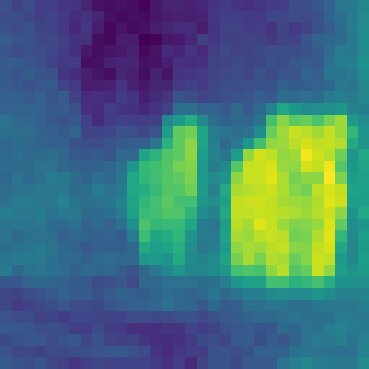}%
        \attnbottombear{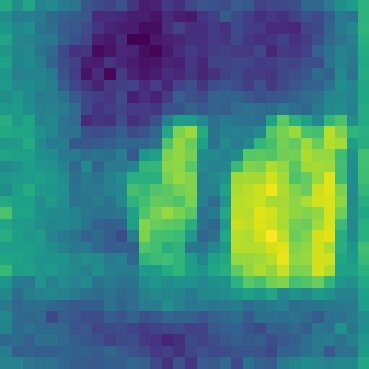}%
        \attnbottombear{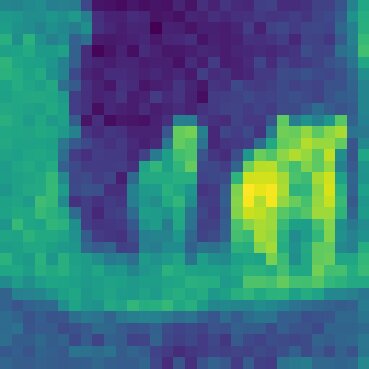}%
    }
   \caption{\textbf{Attention maps for the subject \textit{``bear''}.} Extracted from diffusion timestep 13 of 28 across all 24 DiT blocks for the prompt \textit{``A horse and a bear in a forest''}, using Stable Diffusion 3.5 with JEDI. The final generated image is shown in \cref{fig:intro_example}.}
  \label{fig:attn_bear}
\end{figure*}

\newcommand{\attntophorse}[1]{
    \begin{subfigure}[t]{0.07\linewidth}
        \captionsetup{labelformat=empty}
        \caption{\the\numexpr #1 + 1\relax.}%
        \includegraphics[width=\linewidth]{figures/attn_viz/attn_horse#1.jpg}%
    \end{subfigure}
}

\newcommand{\attnbottomhorse}[1]{
    \begin{subfigure}[t]{0.07\linewidth}
        \captionsetup{labelformat=empty}
        \includegraphics[width=\linewidth]{figures/attn_viz/attn_horse#1.jpg}%
        \caption{\the\numexpr #1 + 1\relax.}%
    \end{subfigure}
}

\begin{figure*}
    \makebox[\linewidth][c]{%
        \attntophorse{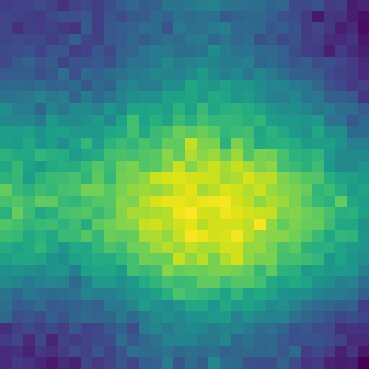}%
        \attntophorse{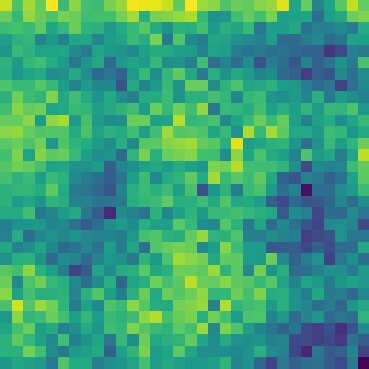}%
        \attntophorse{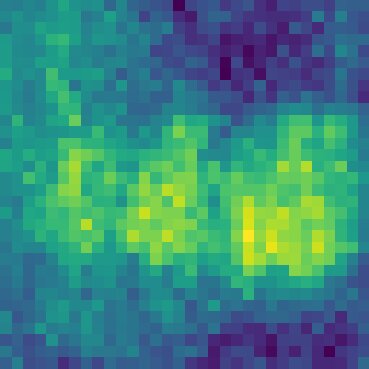}%
        \attntophorse{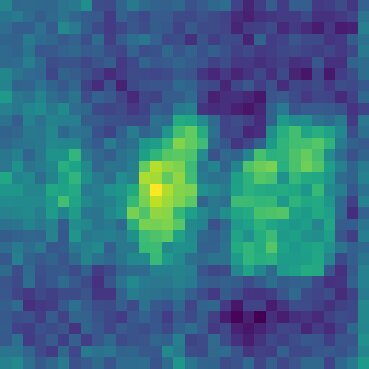}%
        \attntophorse{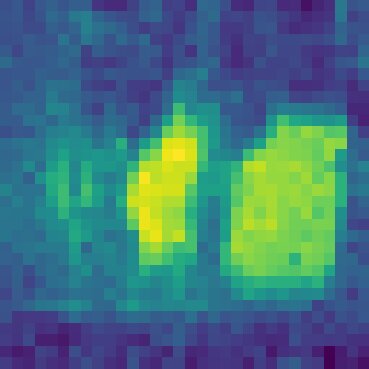}%
        \attntophorse{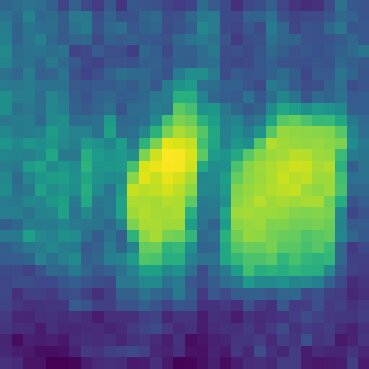}%
        \attntophorse{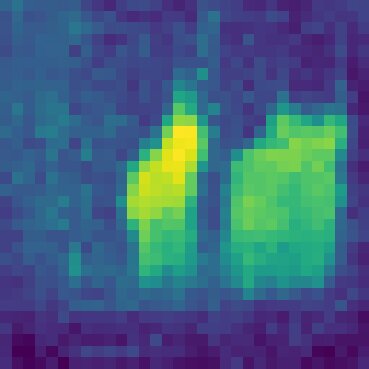}%
        \attntophorse{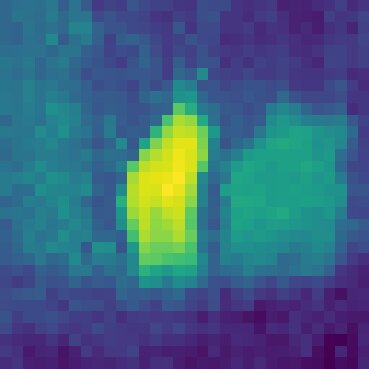}%
        \attntophorse{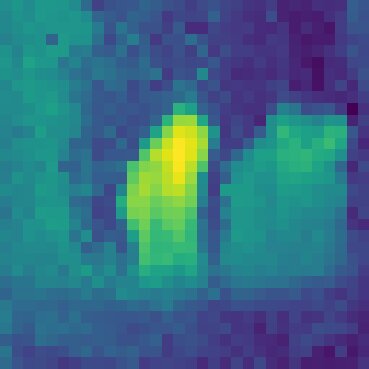}%
        \attntophorse{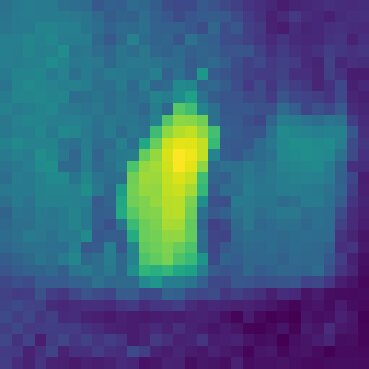}%
        \attntophorse{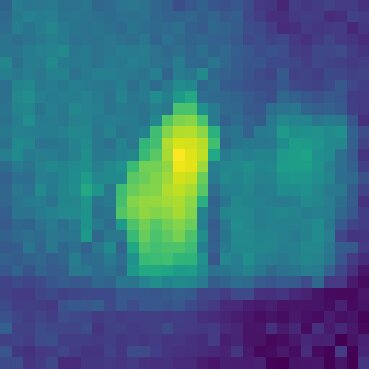}%
        \attntophorse{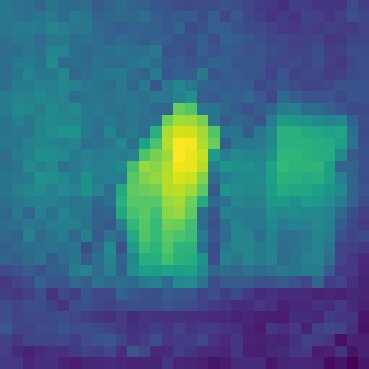}%
    }
    \makebox[\linewidth][c]{%
        \attnbottomhorse{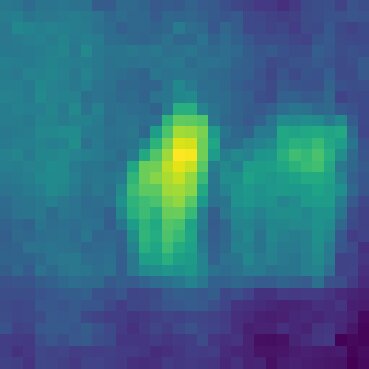}%
        \attnbottomhorse{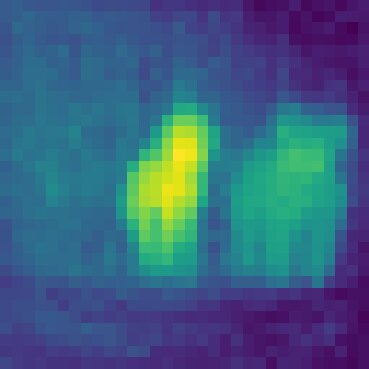}%
        \attnbottomhorse{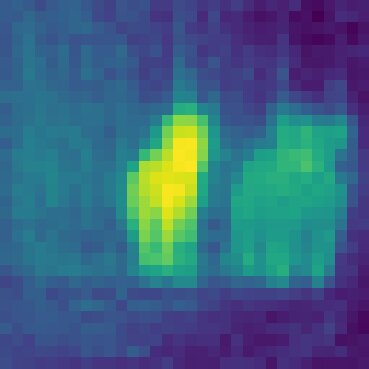}%
        \attnbottomhorse{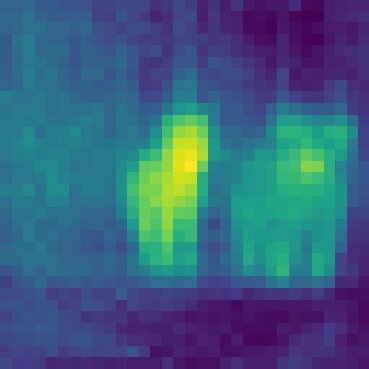}%
        \attnbottomhorse{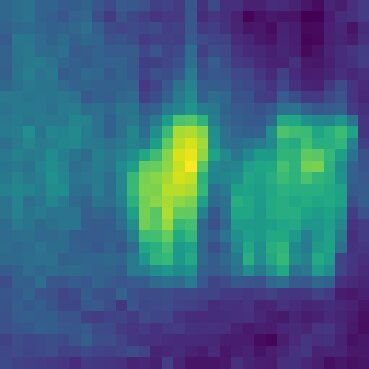}%
        \attnbottomhorse{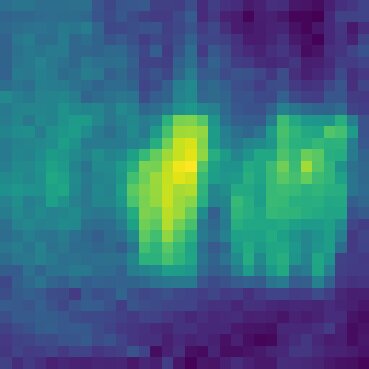}%
        \attnbottomhorse{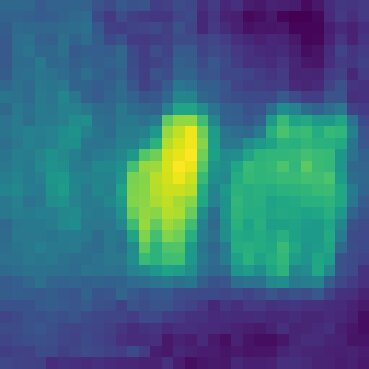}%
        \attnbottomhorse{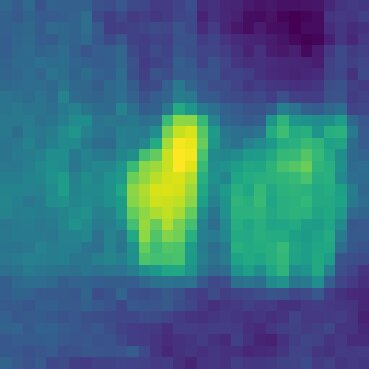}%
        \attnbottomhorse{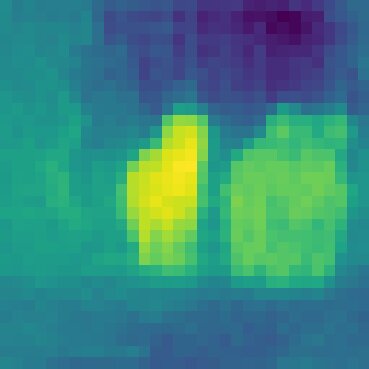}%
        \attnbottomhorse{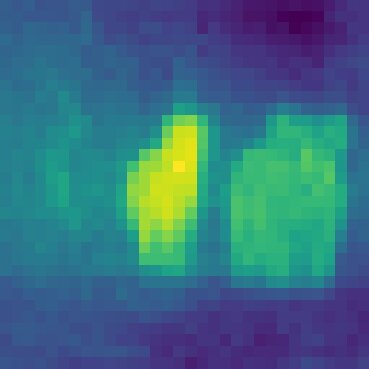}%
        \attnbottomhorse{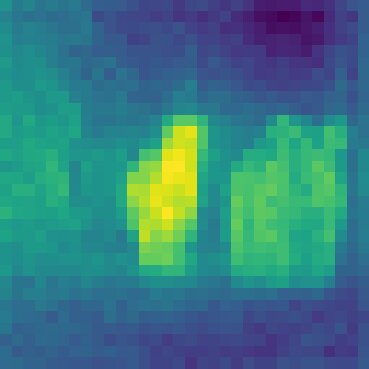}%
        \attnbottomhorse{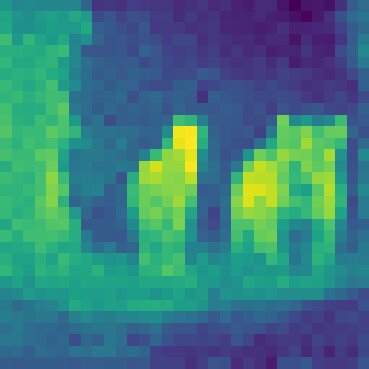}%
    }
   \caption{\textbf{Attention maps for the subject \textit{``horse''}.} Extracted from diffusion timestep 13 of 28 across all 24 DiT blocks for the prompt \textit{``A horse and a bear in a forest''}, using Stable Diffusion 3.5 with JEDI. The final generated image is shown in \cref{fig:intro_example}.}
   \label{fig:attn_horse}
\end{figure*}

\section{Score}
\cref{fig:jsd_evol} presents the inter-group JSD between the two subjects from \cref{fig:intro_example}, computed across DiT blocks 7 to 15 over all diffusion timesteps. The image without attribute mixing exhibits consistently higher inter-group JSD values from timestep 5 onward, indicating stronger subject disentanglement.

\begin{figure}[H]
  \centering
  \begin{tikzpicture}
    \begin{axis}[
      width=\linewidth,
      height=0.6\linewidth,
      xlabel={Timestep},
      ylabel={Intra-Group JSD},
      ymin=0, ymax=0.8,
      xmin=0, xmax=27,
      legend pos=north east,
      axis on top,
      grid=major
    ]

    \addplot[blue, opacity=0.5] table [col sep=comma, x=timestep, y=jedi_0] {data/jsd_data.csv};
    \addplot[blue, opacity=0.5] table [col sep=comma, x=timestep, y=jedi_1] {data/jsd_data.csv};
    \addplot[blue, opacity=0.5] table [col sep=comma, x=timestep, y=jedi_2] {data/jsd_data.csv};
    \addplot[blue, opacity=0.5] table [col sep=comma, x=timestep, y=jedi_3] {data/jsd_data.csv};
    \addplot[blue, opacity=0.5] table [col sep=comma, x=timestep, y=jedi_4] {data/jsd_data.csv};
    \addplot[blue, opacity=0.5] table [col sep=comma, x=timestep, y=jedi_5] {data/jsd_data.csv};
    \addplot[blue, opacity=0.5] table [col sep=comma, x=timestep, y=jedi_6] {data/jsd_data.csv};
    \addplot[blue, opacity=0.5] table [col sep=comma, x=timestep, y=jedi_7] {data/jsd_data.csv};
    \addplot[blue, opacity=0.5] table [col sep=comma, x=timestep, y=jedi_8] {data/jsd_data.csv};

    \addplot[red, opacity=0.5] table [col sep=comma, x=timestep, y=base_0] {data/jsd_data.csv};
    \addplot[red, opacity=0.5] table [col sep=comma, x=timestep, y=base_1] {data/jsd_data.csv};
    \addplot[red, opacity=0.5] table [col sep=comma, x=timestep, y=base_2] {data/jsd_data.csv};
    \addplot[red, opacity=0.5] table [col sep=comma, x=timestep, y=base_3] {data/jsd_data.csv};
    \addplot[red, opacity=0.5] table [col sep=comma, x=timestep, y=base_4] {data/jsd_data.csv};
    \addplot[red, opacity=0.5] table [col sep=comma, x=timestep, y=base_5] {data/jsd_data.csv};
    \addplot[red, opacity=0.5] table [col sep=comma, x=timestep, y=base_6] {data/jsd_data.csv};
    \addplot[red, opacity=0.5] table [col sep=comma, x=timestep, y=base_7] {data/jsd_data.csv};
    \addplot[red, opacity=0.5] table [col sep=comma, x=timestep, y=base_8] {data/jsd_data.csv};

    \addplot[blue, thick] table [col sep=comma, x=timestep, y=jedi_mean] {data/jsd_data.csv};
    \addplot[red, thick] table [col sep=comma, x=timestep, y=base_mean] {data/jsd_data.csv};
    \addplot [black, dashed, opacity=0.7] coordinates {(17, 0) (17, 0.8)};
    \end{axis}
    \end{tikzpicture}
    \caption{\textbf{Inter-group JSD across diffusion timesteps for the base model (red) and JEDI (blue).} Thick lines show the mean JSD across blocks. JEDI is applied only during the first 18 timesteps, indicated by the dashed vertical line.}
    \label{fig:jsd_evol}
\end{figure}

\section{Samples}\label{sec:samples}
We present additional qualitative results and highlight notable behaviors across different model variants.

\noindent\textbf{Stable Diffusion 3.5.}  
\cref{fig:extra_samples_sd35_1,fig:extra_samples_sd35_2} show samples across a broader range of object categories. A noteworthy observation is that when subjects are already well disentangled (i.e., no visible attribute mixing), JEDI leaves the image unchanged. This occurs because the JEDI loss approaches zero in such cases. For example, see the image pair with \textit{``dachshund''} and \textit{``corgi''}.

\noindent\textbf{Stable Diffusion 1.5.}  
Additional samples are shown in \cref{fig:conform_extended}. Due to CONFORM’s relatively high learning rate, generated images occasionally deviate from the base model’s distribution. For instance, a \textit{``violin''} may appear with an unnatural blue color. This phenomenon is absent in JEDI, as the choice of $\alpha = 3 \times 10^{-3}$ prevents excessive deviation from the base model.

\begin{figure*}[t]
    \centering
    \begin{subfigure}[b]{0.4\textwidth}
        \centering
         {\textit{``\textbf{\texttt{\char`\<}Gosling\texttt{\char`\>}} and \textbf{\texttt{\char`\<}Margot\texttt{\char`\>}} on the Moon''}}\\
        \begin{subfigure}[t]{0.48\textwidth}
            \centering
            \includegraphics[width=\linewidth]{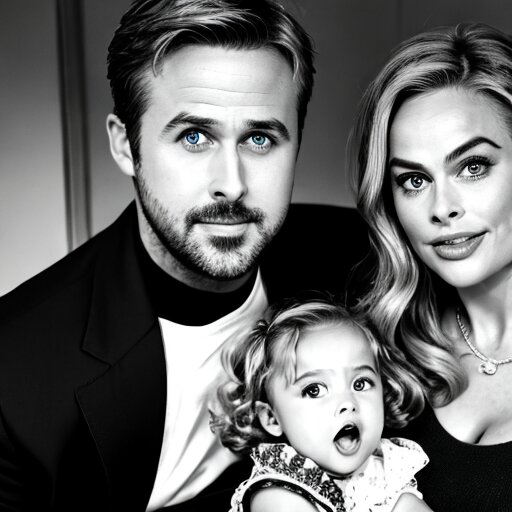}
        \end{subfigure}
        \hfill
        \begin{subfigure}[t]{0.48\textwidth}
            \centering
            \includegraphics[width=\linewidth]{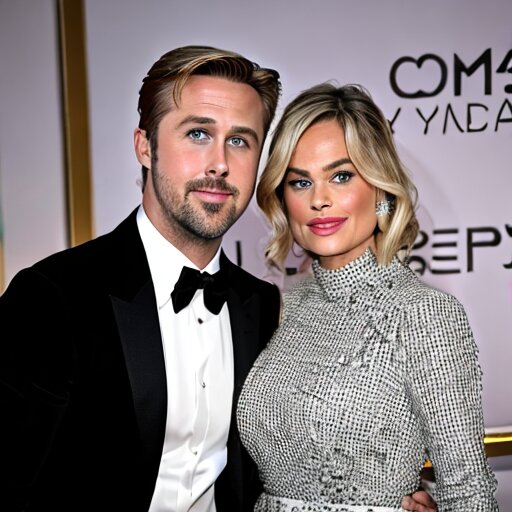}
        \end{subfigure}
    \end{subfigure}
    \hspace{8mm}
    \begin{subfigure}[b]{0.4\textwidth}
        \centering
        {\textit{``\textbf{\texttt{\char`\<}Margot\texttt{\char`\>}} and \textbf{\texttt{\char`\<}Gosling\texttt{\char`\>}} in Times Square''}}\\
        \begin{subfigure}[t]{0.48\textwidth}
            \centering
            \includegraphics[width=\linewidth]{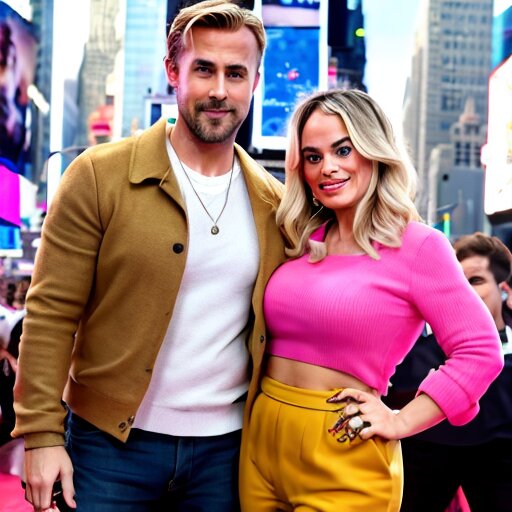}
        \end{subfigure}
        \hfill
        \begin{subfigure}[t]{0.48\textwidth}
            \centering
            \includegraphics[width=\linewidth]{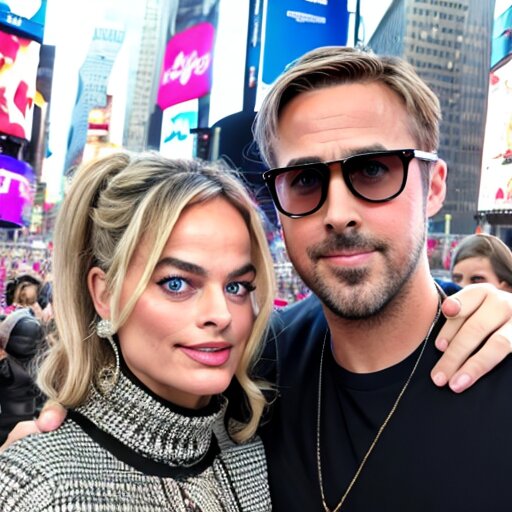}
        \end{subfigure}
    \end{subfigure}\\
    
    \vspace{6mm}

    \centering
     \begin{subfigure}[b]{0.4\textwidth}
        \centering
        {\textit{``\textbf{\texttt{\char`\<}Pitt\texttt{\char`\>}} and \textbf{\texttt{\char`\<}Taylor\texttt{\char`\>}}\\in Venice on a gondola''}}\\
        \begin{subfigure}[t]{0.48\textwidth}
            \centering
            \includegraphics[width=\linewidth]{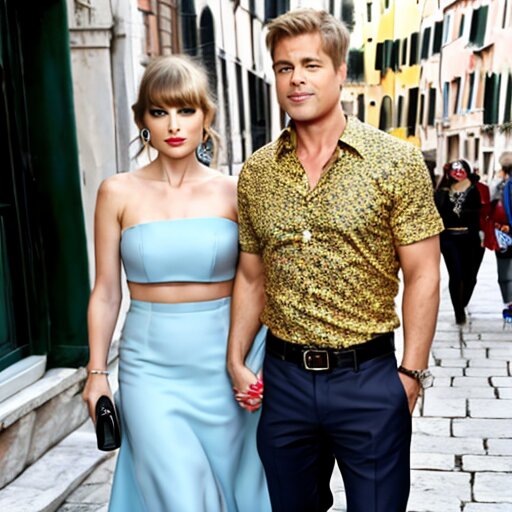}
        \end{subfigure}
        \hfill
        \begin{subfigure}[t]{0.48\textwidth}
            \centering
            \includegraphics[width=\linewidth]{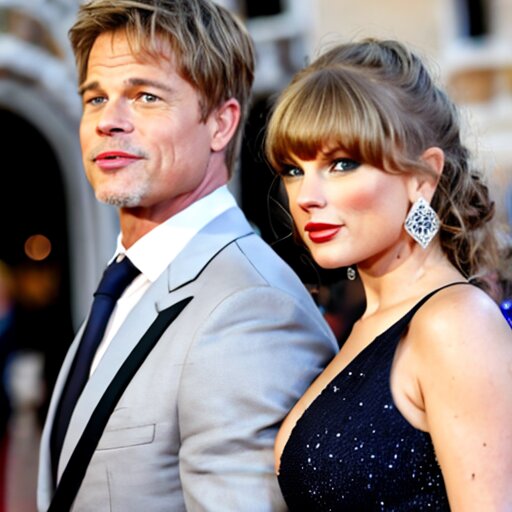}
        \end{subfigure}
    \end{subfigure}
    \hspace{8mm}
    \begin{subfigure}[b]{0.4\textwidth}
        \centering
        {\textit{``\textbf{\texttt{\char`\<}Messi\texttt{\char`\>}} and \textbf{\texttt{\char`\<}Taylor\texttt{\char`\>}} in\\front of Mount Fuji''}}\\
        \begin{subfigure}[t]{0.48\textwidth}
            \centering
            \includegraphics[width=\linewidth]{figures/loraclr/loraclr_without.jpg}
        \end{subfigure}
        \hfill
        \begin{subfigure}[t]{0.48\textwidth}
            \centering
            \includegraphics[width=\linewidth]{figures/loraclr/loraclr_with.jpg}
        \end{subfigure}
    \end{subfigure}\\

    \vspace{6mm}

    \begin{subfigure}[b]{0.4\textwidth}
        \centering
        {\textit{``\textbf{\texttt{\char`\<}LeBron\texttt{\char`\>}} and \textbf{\texttt{\char`\<}Margot\texttt{\char`\>}} at\\the Pyramids of Giza''}} \\
        \begin{subfigure}[t]{0.48\textwidth}
            \centering
            \includegraphics[width=\linewidth]{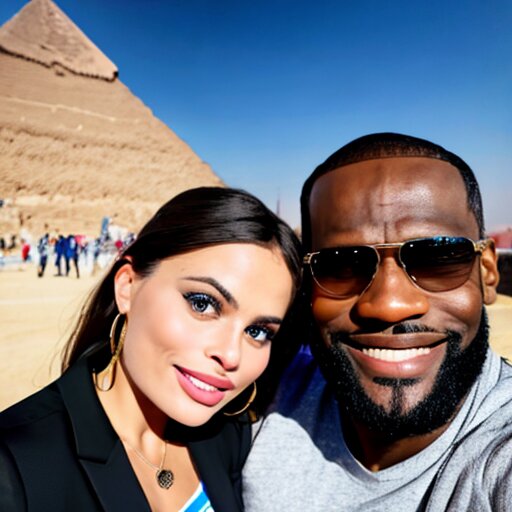}
        \end{subfigure}
        \hfill
        \begin{subfigure}[t]{0.48\textwidth}
            \centering
            \includegraphics[width=\linewidth]{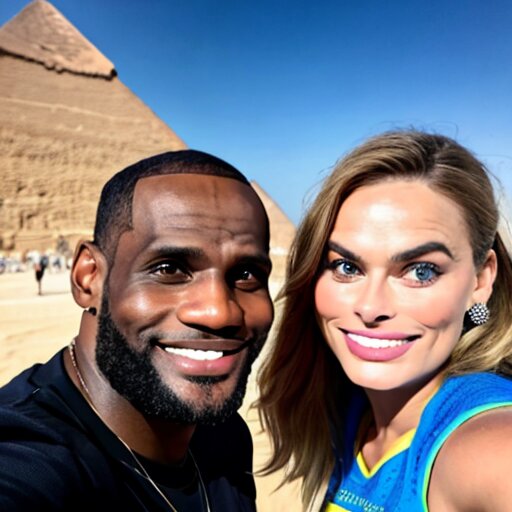}
        \end{subfigure}
    \end{subfigure}
    \hspace{8mm}
    \begin{subfigure}[b]{0.4\textwidth}
        \centering
        {\textit{``\textbf{\texttt{\char`\<}Messi\texttt{\char`\>}} and \textbf{\texttt{\char`\<}Taylor\texttt{\char`\>}} on\\the Great Wall of China''}}\\
        \begin{subfigure}[t]{0.48\textwidth}
            \centering
            \includegraphics[width=\linewidth]{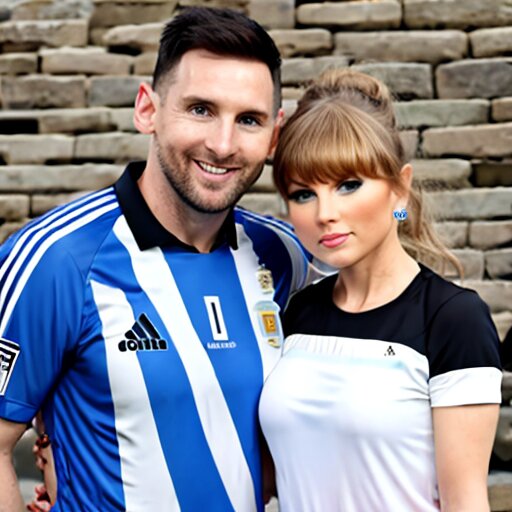}
        \end{subfigure}
        \hfill
        \begin{subfigure}[t]{0.48\textwidth}
            \centering
            \includegraphics[width=\linewidth]{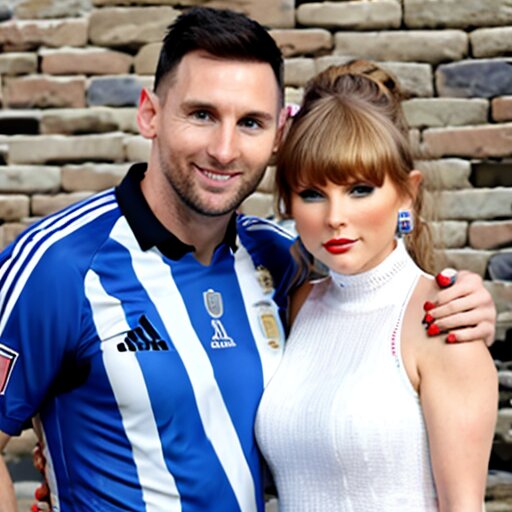}
        \end{subfigure}
    \end{subfigure}
    
    \vspace{6mm}

    \begin{subfigure}[b]{0.4\textwidth}
        \centering
        {\textit{``\textbf{\texttt{\char`\<}LeBron\texttt{\char`\>}} and \textbf{\texttt{\char`\<}Messi\texttt{\char`\>}} at\\the Tokyo Shibuya Crossing''}} \\
        \begin{subfigure}[t]{0.48\textwidth}
            \centering
            \includegraphics[width=\linewidth]{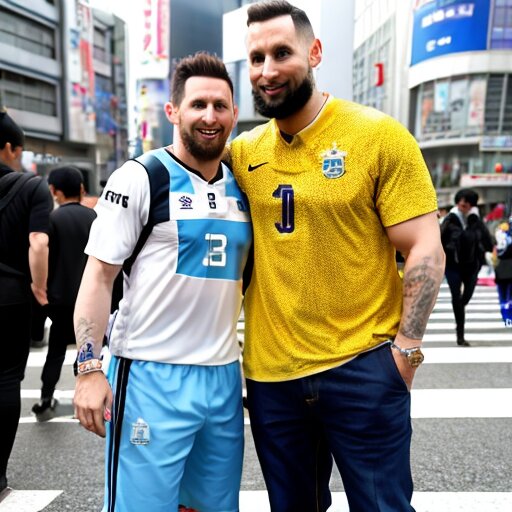}
        \end{subfigure}
        \hfill
        \begin{subfigure}[t]{0.48\textwidth}
            \centering
            \includegraphics[width=\linewidth]{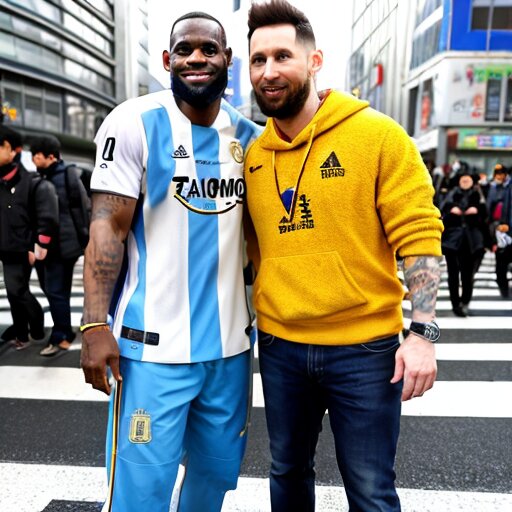}
        \end{subfigure}
    \end{subfigure}
    \hspace{8mm}
    \begin{subfigure}[b]{0.4\textwidth}
        \centering
        {\textit{``\textbf{\texttt{\char`\<}Margot\texttt{\char`\>}} and \textbf{\texttt{\char`\<}Pitt\texttt{\char`\>}} at\\on a ski lift in the Swiss Alps''}}\\
        \begin{subfigure}[t]{0.48\textwidth}
            \centering
            \includegraphics[width=\linewidth]{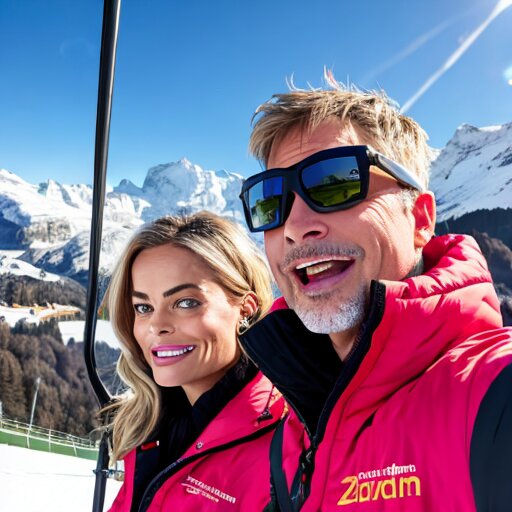}
        \end{subfigure}
        \hfill
        \begin{subfigure}[t]{0.48\textwidth}
            \centering
            \includegraphics[width=\linewidth]{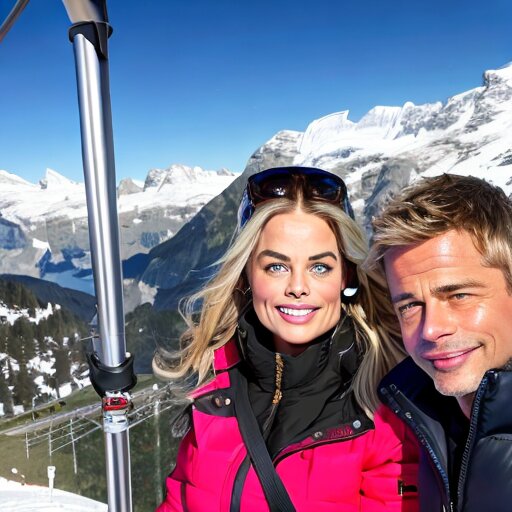}
        \end{subfigure}
    \end{subfigure}
    \caption{\textbf{Comparison between LoRACLR (left) and LoRACLR + JEDI (right).} The baseline model exhibits attribute mixing between subjects (e.g., ``Taylor'' appears in football attire), whereas LoRACLR + JEDI achieves clearer subject disentanglement and preserves subject-specific features.}
    \label{fig:loraclr_samples}
\end{figure*}

\noindent\textbf{LoRACLR.}  
Further LoRACLR results are shown in \cref{fig:loraclr_samples}. Since the LoRACLR model combines 14 concepts---many involving famous figures from film or sports---it occasionally disregards background details, leading to misalignment between the prompt and the image. However, this issue is inherent to the LoRACLR model and not introduced by JEDI. We solely demonstrate that JEDI can successfully disentangle the subjects of the prompt.

\newcommand{\sixImageRow}[5]{
    \begin{subfigure}[t]{\textwidth}
        \centering
        \textit{#1}\\
        \begin{subfigure}[t]{0.155\textwidth}
            \centering
            \includegraphics[width=\linewidth]{figures/pairs/#2/base/image_#3.jpg}
        \end{subfigure}
        \begin{subfigure}[t]{0.155\textwidth}
            \centering
            \includegraphics[width=\linewidth]{figures/pairs/#2/jsd/image_#3.jpg}
        \end{subfigure}
        \hfill
        \begin{subfigure}[t]{0.155\textwidth}
            \centering
            \includegraphics[width=\linewidth]{figures/pairs/#2/base/image_#4.jpg}
        \end{subfigure}
        \begin{subfigure}[t]{0.155\textwidth}
            \centering
            \includegraphics[width=\linewidth]{figures/pairs/#2/jsd/image_#4.jpg}
        \end{subfigure}
        \hfill
        \begin{subfigure}[t]{0.155\textwidth}
            \centering
            \includegraphics[width=\linewidth]{figures/pairs/#2/base/image_#5.jpg}
        \end{subfigure}
        \begin{subfigure}[t]{0.155\textwidth}
            \centering
            \includegraphics[width=\linewidth]{figures/pairs/#2/jsd/image_#5.jpg}
        \end{subfigure}\\
    \end{subfigure}
    \\\vspace{1em}
}

\begin{figure*}[t]
    \centering
    \sixImageRow{``A \textbf{horse} and a \textbf{bear} in a forest''}{bear_horse}{0}{12}{36}
    \sixImageRow{``A \textbf{sparrow} and a \textbf{finch} perched on a blossoming branch''}{sparrow}{16}{68}{81}
    \sixImageRow{``An \textbf{apple} and a \textbf{pear} hanging from adjacent branches in an orchard''}{apple_pear}{19}{67}{79}
    \sixImageRow{``A \textbf{dachshund} and a \textbf{corgi} sitting together on a cozy rug''}{dachshund_corgi}{2}{54}{80}
    \sixImageRow{``A \textbf{canoe} and a \textbf{kayak} tied to a wooden dock at dawn''}{canoe}{5}{31}{44}
    \sixImageRow{``A \textbf{street bike} and a \textbf{dirt bike} leaning against a garage wall''}{streetbike}{8}{34}{60}
    \caption{\textbf{Side-by-side comparison of Stable Diffusion 3.5 (left) and Stable Diffusion 3.5 + JEDI (right).} Each image pair was generated under identical conditions with a guidance scale of $4.5$ and $28$ inference steps.}
    \label{fig:extra_samples_sd35_1}
\end{figure*}

\begin{figure*}[t]
    \centering
    \sixImageRow{``A \textbf{sailboat} and a \textbf{yacht} anchored in a calm harbor at sunset''}{sailboat}{10}{88}{114}
    \sixImageRow{``A \textbf{sheep} and a \textbf{goat} grazing in a misty pasture''}{sheep_goat}{4}{17}{95}
    \sixImageRow{``A \textbf{rabbit} and a \textbf{hare} nibbling grass in a sunlit meadow''}{rabbit_hare}{47}{95}{119}
    \sixImageRow{``A \textbf{dolphin} and a \textbf{whale} breaching near each other in the ocean''}{dolphin_whale}{34}{58}{70}
    \sixImageRow{``A \textbf{maple leaf} and an \textbf{oak leaf} lying on a forest floor covered in moss''}{maple_leaf}{20}{32}{68}
    \sixImageRow{``A \textbf{jaguar} and a \textbf{leopard} crouching in dense rainforest foliage''}{jaguar}{17}{65}{12}

    \caption{\textbf{Side-by-side comparison of Stable Diffusion 3.5 (left) and Stable Diffusion 3.5 + JEDI (right).} Each image pair was generated under identical conditions with a guidance scale of $4.5$ and $28$ inference steps.}
    \label{fig:extra_samples_sd35_2}
\end{figure*}
\newcommand{\siximageblock}[4]{%
    \begin{subfigure}[t]{\linewidth}
        \centering
        \begin{subfigure}[t]{0.155\linewidth}
            \centering
            \captionsetup{labelformat=empty}
            \ifthenelse{\equal{#4}{true}}{\caption{\textbf{SD 1.5}}}{}
            \includegraphics[width=\linewidth]{figures/sd1.5/base/image_#1.jpg}
        \end{subfigure}
        \begin{subfigure}[t]{0.155\linewidth}
            \centering
            \captionsetup{labelformat=empty}
            \ifthenelse{\equal{#4}{true}}{\caption{\textbf{JEDI (ours)}}}{}
            \includegraphics[width=\linewidth]{figures/sd1.5/jedi/image_#1.jpg}
        \end{subfigure}
        \begin{subfigure}[t]{0.155\linewidth}
            \centering
            \captionsetup{labelformat=empty}
            \ifthenelse{\equal{#4}{true}}{\caption{\textbf{CONFORM}}}{}
            \includegraphics[width=\linewidth]{figures/sd1.5/conform/image_#1.jpg}
        \end{subfigure}
        \hfill
        \begin{subfigure}[t]{0.155\linewidth}
            \centering
            \captionsetup{labelformat=empty}
            \ifthenelse{\equal{#4}{true}}{\caption{\textbf{SD 1.5}}}{}
            \includegraphics[width=\linewidth]{figures/sd1.5/base/image_#2.jpg}
        \end{subfigure}
        \begin{subfigure}[t]{0.155\linewidth}
            \centering
            \captionsetup{labelformat=empty}
            \ifthenelse{\equal{#4}{true}}{\caption{\textbf{JEDI (ours)}}}{}
            \includegraphics[width=\linewidth]{figures/sd1.5/jedi/image_#2.jpg}
        \end{subfigure}
        \begin{subfigure}[t]{0.155\linewidth}
            \centering
            \captionsetup{labelformat=empty}
            \ifthenelse{\equal{#4}{true}}{\caption{\textbf{CONFORM}}}{}
            \includegraphics[width=\linewidth]{figures/sd1.5/conform/image_#2.jpg}
        \end{subfigure}\\\vspace{1mm}
        \textit{#3}
        \vspace{2mm}
    \end{subfigure}%
}

\begin{figure*}
    \siximageblock{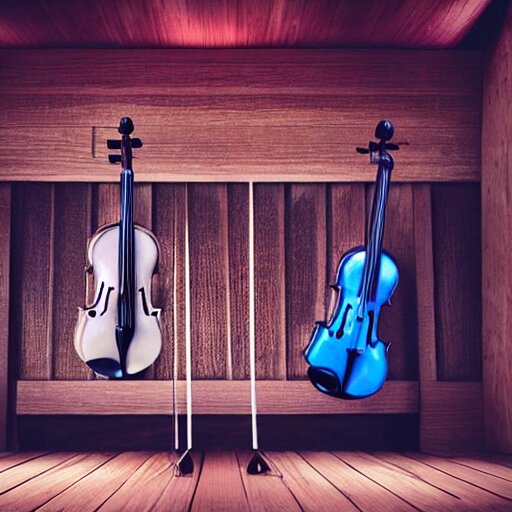}{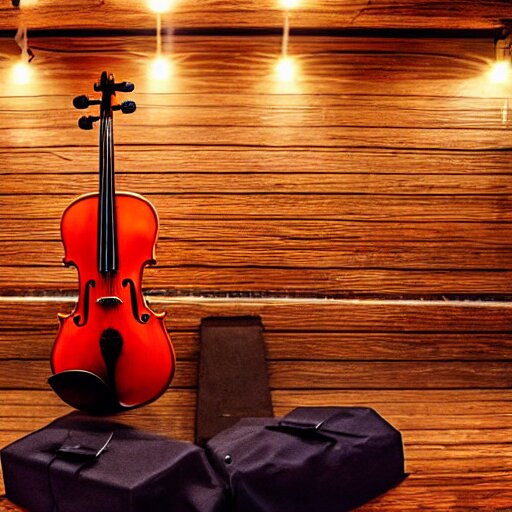}{``A \textbf{violin} and a \textbf{viola} on a wooden stage under soft spotlights''}{true}
    \siximageblock{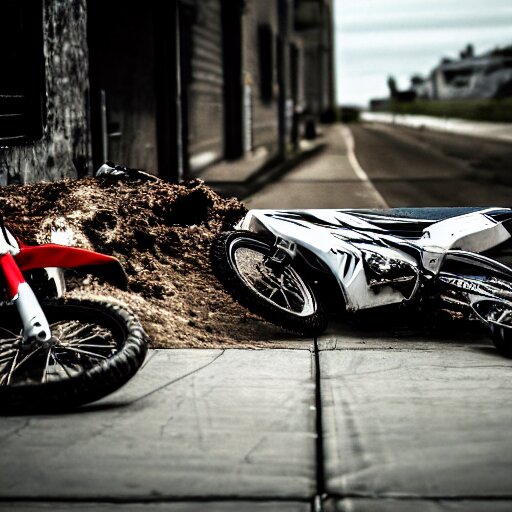}{109}{``A \textbf{street bike} and a \textbf{dirt bike} leaning against a garage wall''}{false}
    \siximageblock{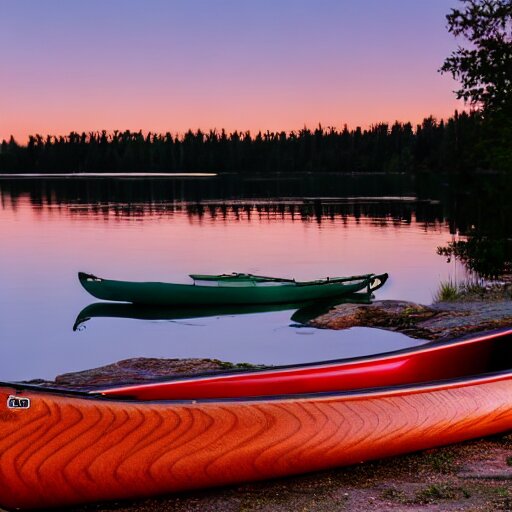}{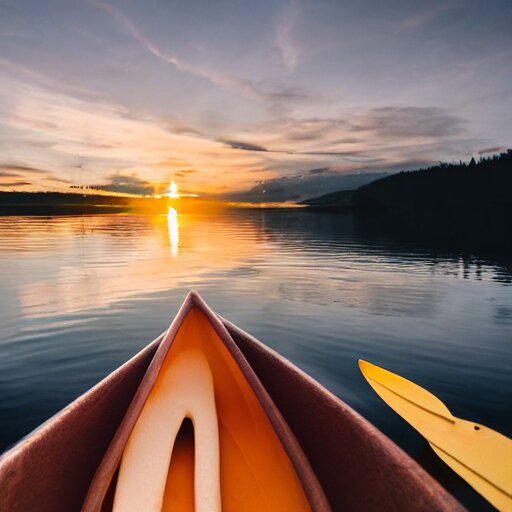}{``A \textbf{canoe} and a \textbf{kayak} tied to a wooden dock at dawn''}{false}
    \siximageblock{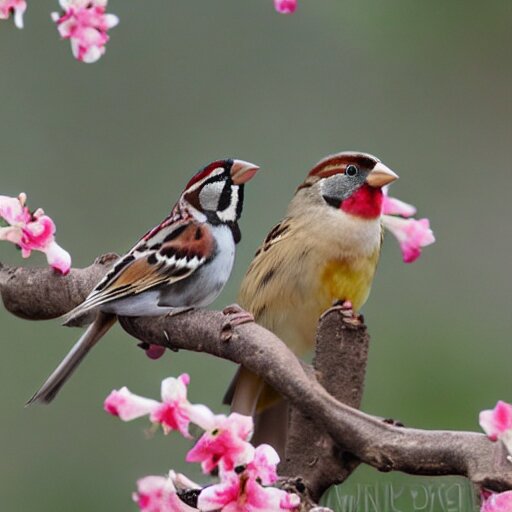}{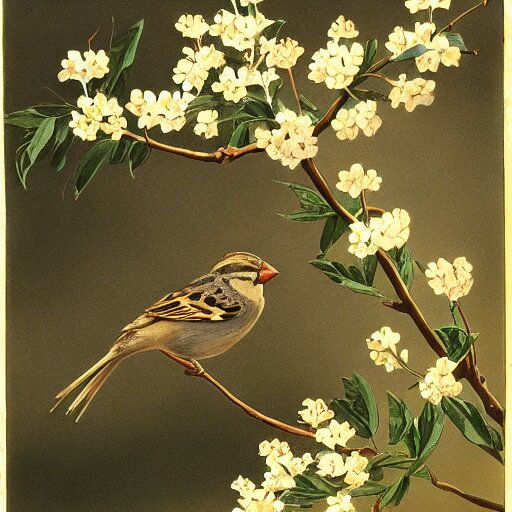}{``A \textbf{sparrow} and a \textbf{finch} perched on a blossoming branch''}{false}
    \siximageblock{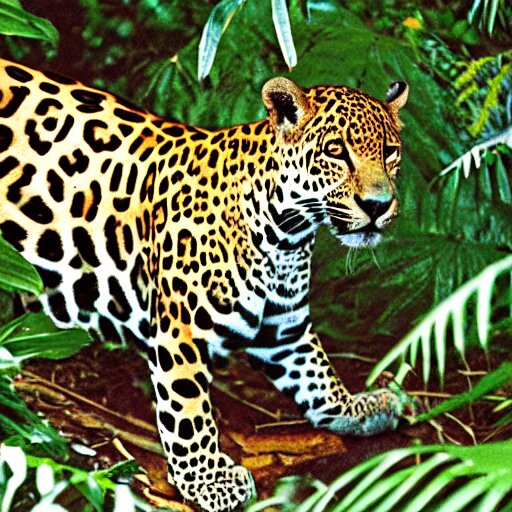}{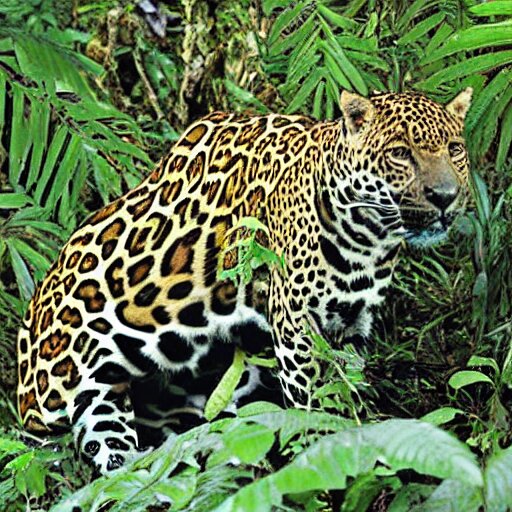}{``A \textbf{jaguar} and a \textbf{leopard} crouching in dense rainforest foliage''}{false}
    \siximageblock{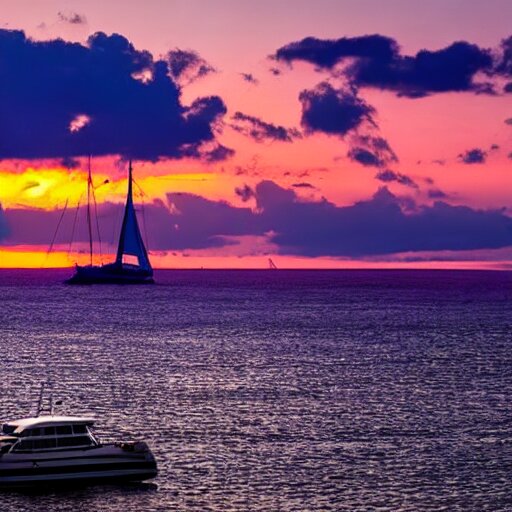}{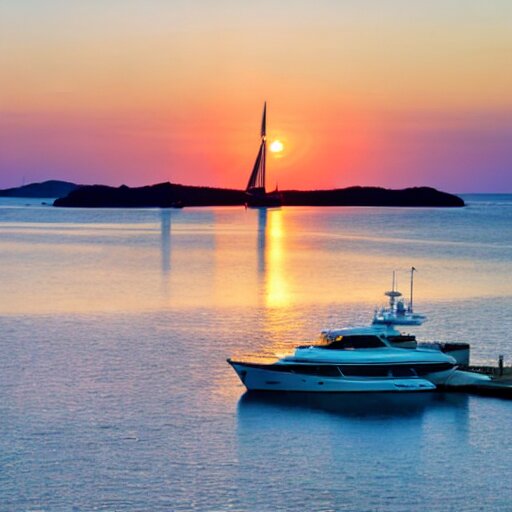}{``A \textbf{sailboat} and a \textbf{yacht} anchored in a calm harbor at sunset''}{false}
    \vspace{-5mm}
    \caption{\textbf{Comparison of JEDI and CONFORM on Stable Diffusion 1.5.} Each image triplet was generated under identical conditions with 50 inference steps and a guidance scale of 7.5. For details of each method, refer to \cref{sec:implementation}.}
    \label{fig:conform_extended}
\end{figure*}

\end{document}